%% file: acl_latex.tex
\pdfoutput=1

\documentclass[11pt]{article}

\usepackage[preprint]{acl}
\usepackage{times}
\usepackage{latexsym}
\usepackage{booktabs}
\usepackage{multicol}
\usepackage{multirow}
\usepackage[T1]{fontenc}
\usepackage[utf8]{inputenc}
\usepackage{microtype}
\usepackage{inconsolata}
\usepackage{graphicx}
\usepackage{xspace}
\usepackage{amsmath, amssymb}
\usepackage{enumitem}
\usepackage{xcolor}
\usepackage{cleveref}

\usepackage{multirow}
\usepackage{longtable}
\usepackage{listings}
\usepackage{graphicx}
\usepackage{mathtools}
\usepackage{amsthm}
\usepackage{xspace}
\usepackage[subtle]{savetrees}
\usepackage[most]{tcolorbox}
\usepackage{colortbl}
\usepackage{bbm}
\usepackage{float}
\usepackage{array}

\newcommand{\OURDATA}{\texttt{SubPOP}\xspace}

\definecolor{berkeleyblue}{HTML}{3B7EA1}
\definecolor{berkeleygold}{HTML}{FDB515}
\definecolor{lightcyan}{HTML}{E5F2F2}
\definecolor{main}{HTML}{4472C4}   
\definecolor{sub}{HTML}{EBF4FF}

\definecolor{lightcyan}{rgb}{0.88,1,1}

\newcommand{\highlightrow}{\rowcolor{berkeleygold!20}}
\newcommand{\highlightrowtwo}{\rowcolor{berkeleyblue!20}}

\DeclareMathSizes{10}{9}{6}{5}

\title{Language Model Fine-Tuning on Scaled Survey Data\\ for Predicting Distributions of Public Opinions}

\author{
 \textbf{Joseph Suh\textsuperscript{1}$^{*}$},
 \textbf{Erfan Jahanparast\textsuperscript{1}$^{*}$},
 \textbf{Suhong Moon\textsuperscript{1}$^{*}$},
 \textbf{Minwoo Kang\textsuperscript{1}$^{*}$},
 \textbf{Serina Chang\textsuperscript{1,2}}
\\
\\
 \textsuperscript{1}University of California, Berkeley
 \textsuperscript{2}Microsoft Research
\\
 \small{
   \textbf{Correspondence:} \{josephsuh,serinac\}@berkeley.edu
 }
}

\begin{document}
\maketitle
\def\thefootnote{$*$}\footnotetext[1]{Equal Contribution.}\def\thefootnote{\arabic{footnote}}
\input{sections/s0_abstract}
\input{sections/s1_introduction}

\input{sections/s2_related_work}
\input{sections/s3_method}
\input{sections/s4_experiment}
\input{sections/s5_conclusion}

\section*{Acknowledgments}
We sincerely appreciate Prof. John Canny for constructive comments and guidance. 
We also thank Prof. Emma Pierson and Dr. Sehoon Kim for valuable discussions as well as Woosuk Kwon for his support with vLLM.
J.S. and S.M. would like to acknowledge the support from the Korea Foundation for Advanced Studies (KFAS).
S.M. is supported by BAIR-Google Commons and M.K. is supported by the Apple Ph.D. Fellowship in Integrated Systems.
We gratefully acknowledge the generous support from Strong Compute for providing fully managed GPU for fine tuning through the Strong Compute Instant Super Computer system, including support from Adam Peaston.

\bibliography{references}

\appendix
\input{sections/a1_dataset}
\input{sections/a2_sft}
\input{sections/a3_chat}
\input{sections/a4_baseline_detail}
\input{sections/a5_main_result_finegrained}
\end{document}

%% file: sections/s0_abstract.tex
\begin{abstract}
Large language models (LLMs) present novel opportunities in public opinion research by predicting survey responses in advance during the early stages of survey design.
Prior methods steer LLMs via descriptions of subpopulations as LLMs' input prompt, yet such prompt engineering approaches have struggled to faithfully predict the distribution of survey responses from human subjects.
In this work, we propose directly fine-tuning LLMs to predict response distributions by leveraging unique structural characteristics of survey data.
To enable fine-tuning, we curate \OURDATA, a significantly scaled dataset of 3,362 questions and 70K subpopulation-response pairs from well-established public opinion surveys.
We show that fine-tuning on \OURDATA greatly improves the match between LLM predictions and human responses across various subpopulations, reducing the LLM-human gap by up to 46\% compared to baselines, and achieves strong generalization to unseen surveys and subpopulations.
Our findings highlight the potential of survey-based fine-tuning to improve opinion prediction for diverse, real-world subpopulations and therefore enable more efficient survey designs.
Our code is available at \url{https://github.com/JosephJeesungSuh/subpop}.
\end{abstract}

%% file: sections/s1_introduction.tex
\section{Introduction}
Surveys provide an essential tool for probing public opinions on societal issues, especially as opinions vary over time and across subpopulations.
However, surveys are also costly, time-consuming, and require careful calibration to mitigate non-response and sampling biases \cite{choi2004catalog, bethlehem2010selection}. 
Recent work suggests that large language models (LLMs) can assist public opinion studies by predicting survey responses across different subpopulations, explored in both social science ~\cite{argyle2023out,bail2024can,ashokkumar2024predicting,manning2024automated} and NLP~\cite{santurkar2023whose,chu2023language,moon-etal-2024-virtual,hamalainen2023evaluating,chiang2023can}.
Such capabilities could substantially enhance the survey development process, not as a replacement for human participants but as a 
tool for researchers to conduct pilot testing, identify subpopulations to over-sample, and test analysis pipelines prior to conducting the full survey  \cite{rothschild2024survey}.

\begin{figure}
    \centering
    \includegraphics[width=1.0\linewidth]{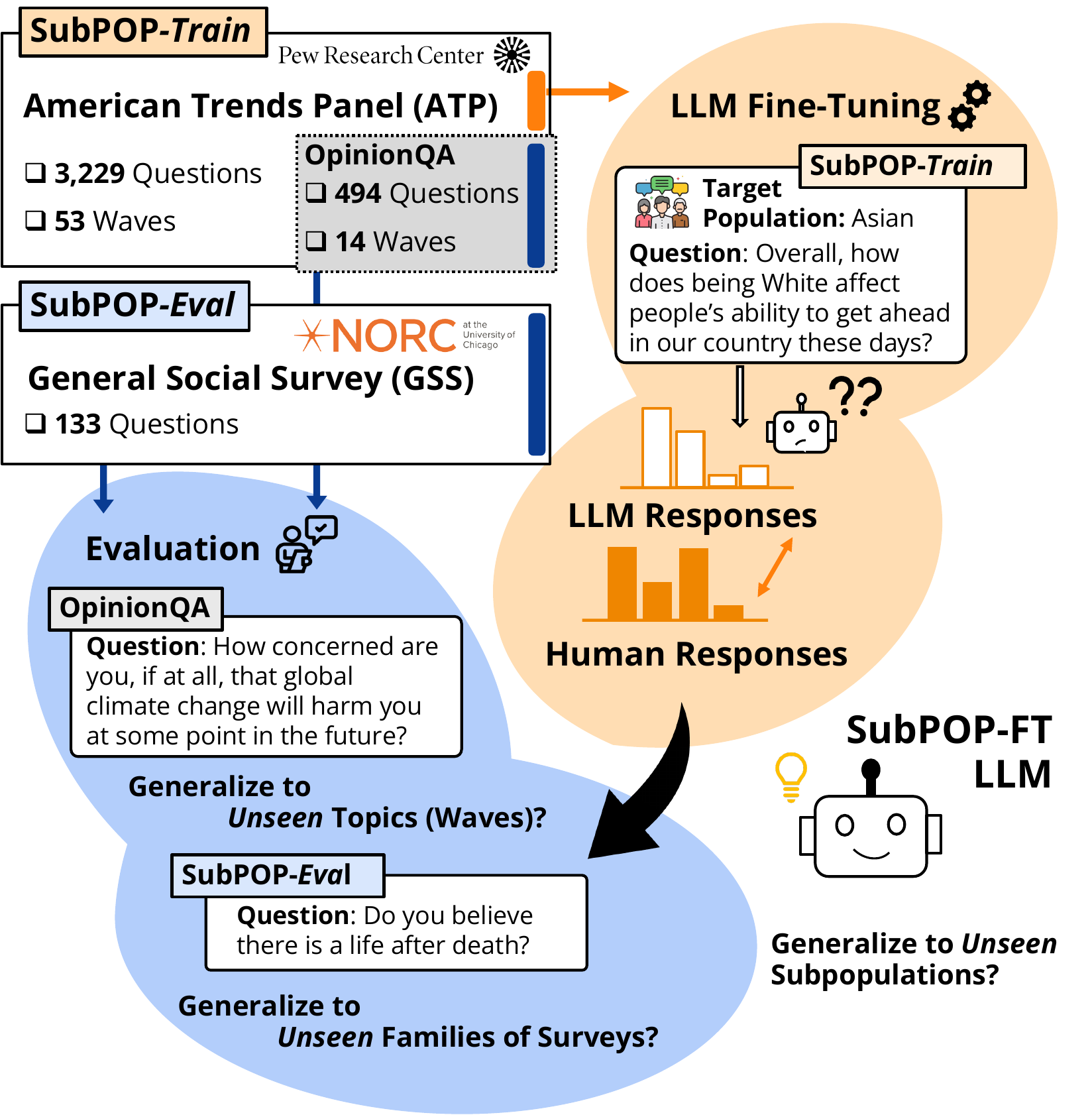}
    \caption{Illustration of our method and \OURDATA. We collect survey data from two survey families—ATP from Pew Research~\cite{atp} (forming \OURDATA-Train) and GSS from NORC~\cite{davern2024gss} (forming \OURDATA-Eval). 
    LLMs are fine-tuned on \OURDATA-Train and evaluated on both OpinionQA~\cite{santurkar2023whose} and \OURDATA-Eval to assess generalization of distributional opinion prediction across unseen survey topics, survey families, and subpopulations.
    }
    \label{fig:teaser}
\end{figure}

Prior work in steering language models, \textit{i.e.} conditioning models to reflect the opinions of a specific subpopulation, has primarily investigated different prompt engineering techniques~\cite{santurkar2023whose, moon-etal-2024-virtual, park2024generative}. However, prompting alone has shown limited success in generating completions that accurately reflect the distributions of survey responses collected from human subjects. Off-the-shelf LLMs~\cite{achiam2023gpt, dubey2024llama, jiang2023mistral} have shown to mirror the opinions of certain US subpopulations such as the wealthy and educated \cite{santurkar2023whose,gallegos2024bias,deshpande2023toxicity,kim2023ai}, while generating stereotypical or biased predictions of underrepresented groups ~\cite{cheng2023compost,cheng2023marked,wang2024large}. Furthermore, these models often fail to capture the variation of human opinions within a subpopulation \cite{kapania2024simulacrum, park2024diminished}.
While fine-tuning presents opportunities to address these limitations ~\cite{chu2023language, he2024community}, existing methods fail to train models that accurately predict opinion distributions across diverse survey question topics and subpopulations.

\vspace{-5pt}
\paragraph{The present work.}
Here, we propose directly fine-tuning LLMs on large-scale, high-quality survey data,
consisting of questions about diverse topics and responses from each subpopulation, defined by demographic, socioeconomic, and ideological traits.
By casting pairs of (subpopulation, survey question) as input prompts, we train the LLM to align its response distribution against that of human subjects in a supervised manner.
We posit that survey data is particularly well-suited for fine-tuning LLMs since: (1) We can train the model with clear \textbf{subpopulation-response pairs} that explicitly link group identities and expressed opinions,
which is rare in LLMs' pre-training corpora,
(2) Large-scale opinion polls are carefully designed and calibrated (\textit{e.g.} using post-stratification) to estimate \textbf{representative} human responses, in contrast with LLMs' pre-training data where certain populations are over- or underrepresented, 
(3) Our training objective explicitly aligns model predictions with response \textbf{distributions} from each subpopulation, enabling LLMs to capture variance within human subpopulations.

Training on public opinion survey data has remained under-explored due to the limited availability of structured survey datasets. 
To this end, we curate and release \textbf{\OURDATA} (\textbf{Sub}population-level \textbf{P}ublic \textbf{O}pinion \textbf{P}rediction), a dataset of 70K subpopulation-response distribution pairs ($6.5\times$ larger compared to previous datasets).
We show that fine-tuning LLMs on \OURDATA significantly improves the distributional match between LLM generated and human responses, and improvements are consistent across subpopulations of varying sizes.
Additionally, the improvement generalizes to \textit{unseen} subpopulations, survey waves (topics), and survey families, \textit{i.e.} surveys administered by different institutions.
Such broad generalization is particularly critical for real-world public opinions research, where practitioners are most in need of synthetic data for survey questions or subpopulations (or both) that they have not tested before.

Our contributions are summarized as follows:
\vspace{-3mm}
\begin{itemize}[leftmargin=3.3mm]
\setlength\itemsep{2pt}

\item We show that training LLMs on response distributions from survey data significantly improves their ability to predict the opinions of subpopulations, reducing the Wasserstein distance between LLM and human distributions by 32-46\% compared to top-performing baselines. (\Cref{section_experiments_prediction_of_opinion_distributions})
\vspace{-1mm}
\item We show that the performance of the fine-tuned LLMs strongly generalizes to out-of-distribution data, including unseen subpopulations, new survey waves, and different survey families. 
(\Cref{section_experiments_prediction_of_opinion_distributions} and \Cref{section_experiments_per_group})
\vspace{-1mm}
\item We release \OURDATA, a curated and pre-processed dataset of public opinion survey results that is $6.5\times$ larger than existing datasets, enabling fine-tuning at scale.
\end{itemize}

%% file: sections/s2_related_work.tex
\section{Related Work}
\vspace{-5pt}
\paragraph{Public opinion datasets.}
Several research institutions conduct large-scale public opinion polls and release data from those surveys.
Important examples include Pew Research Center's American Trends Panel (ATP), which consists of multiple waves of cross-sectional surveys on different topics~\cite{atp}, and General Social Survey (GSS) from the NORC at the University of Chicago~\cite{davern2024gss}.
Existing datasets have curated such data for evaluating LLM-based opinion predictions, including OpinionQA~\cite{santurkar2023whose}, a subset of ATP survey waves containing about 500 questions on contentious social topics.
While OpinionQA is widely used in prior work~\cite{he2024community, zhao2023group, li2023steerability, li2024culturellm}, we find its total number of questions limited in scale for fine-tuning LLMs and instead use this dataset for evaluation.
We further collect an extended set of survey data from ATP waves not included in OpinionQA, as well as from GSS to curate \OURDATA.
Other datasets, such as GlobalOpinionQA \cite{durmus2023towards}---derived from the World Values Survey (WVS) \cite{wvs2022}
and the Pew Global Attitudes Survey \citep{pewresearch2024}---and the PRISM dataset~\cite{kirk2024prism} 
investigates how language models align with opinions from populations across the globe and different cultures.

\vspace{-5pt}
\paragraph{Predicting human opinions with LLMs.}
Prior work has explored various prompt engineering approaches for steering LLM responses: earlier work use rule-based prompts that incorporate demographic profiles of individuals or populations, or few-shot examples of survey question-response \cite{hwang2023aligning, simmons2022moral, santurkar2023whose, dominguez2023questioning}. Recent work explore prompting LLMs with open-ended text, including interview transcripts \cite{park2024generative}, personal narratives \cite{moon-etal-2024-virtual}, or LLM-refined prompts \cite{kim2024few, sun2024persona}.
Our fine-tuning approach is complementary to prompt engineering methods: while prompt engineering seeks to optimize what information is provided to the LLM (while the model is frozen), fine-tuning seeks to optimize how the model utilizes the provided information (while the prompt is frozen).
In this work, we demonstrate that our fine-tuned models exhibit significant improvements in matching the response distributions of humans without requiring elaborate prompt engineering methods.

Other work \cite{chu2023language, he2024community,feng-etal-2024-modular} fine-tune language models on text corpora from specific communities (\textit{e.g.}, Reddit) to infer the most popular response or response distribution for a given survey question. 
While this approach benefits from large-scale and continuously updated text corpora, it struggles with disproportionate representation online and lacks comprehensive coverage of diverse subpopulations. 
A few works have explored directly fine-tuning on public opinion survey data, but in different problem settings from ours.
\citet{li2023steerability} apply collaborative filtering to individual-level responses to learn embeddings for individuals, and 
\citet{zhao2023group} develop a meta-learning framework to predict the opinions of new groups given a small number of in-context examples for that group.
In contrast, our approach does not require individual-level responses and can generalize to unseen groups and survey questions without \textit{any} responses.

A recent work \cite{li2024culturellm} and a work concurrent to ours \cite{cao2025specializing} also explored fine-tuning LLMs on the World Values Survey (WVS) to align the LLM's opinion response with a culture or entire country populations.
In comparison, our work focuses on US surveys, testing whether LLMs can align with finer-grained subpopulations within one country and whether LLMs fine-tuned on one US-representative survey can generalize to another.
However, we note that our proposed method for fine-tuning language models applies to any survey dataset with distributional information about subpopulation responses.

\vspace{-5pt}
\paragraph{Pluralistic alignment of LLMs.}
Recent literature on pluralistic and distributional alignment target a similar yet different problem in fine-tuning LLMs~\cite{chakrabortymaxmin,melnyk2024distributional,poddar2024personalizing,siththaranjan2023distributional,yao2024no,sorensen2024roadmap,lake2024distributional,chen2024pal,jiang2024can}.
While this line of work shares a similar goal as ours in training models to reflect on opinions (and preferences) of diverse subpopulations, most work differ from ours in that they operate in the context of training against \textit{pair-wise} preference orderings between alternative language model completions, extending the Bradley-Terry-Luce model~\cite{rajkumar2014statistical, ouyang2022training, rafailov2024direct} or investigating alternative models to account for diverging preference orderings across populations.
In contrast, our work trains the model to directly predict the opinion distributions of human subpopulations, where accurately matching distributions across a large variety of subpopulations is of paramount interest.
Our work additionally focuses on the particular context of estimating human opinions about societal issues---the objective of public opinion research---which enables relatively straightforward supervised training on openly available, structured survey data as presented by \OURDATA.

%% file: sections/s3_method.tex
\begin{figure}[!t]
    \centering
    \includegraphics[width=1.00\linewidth]{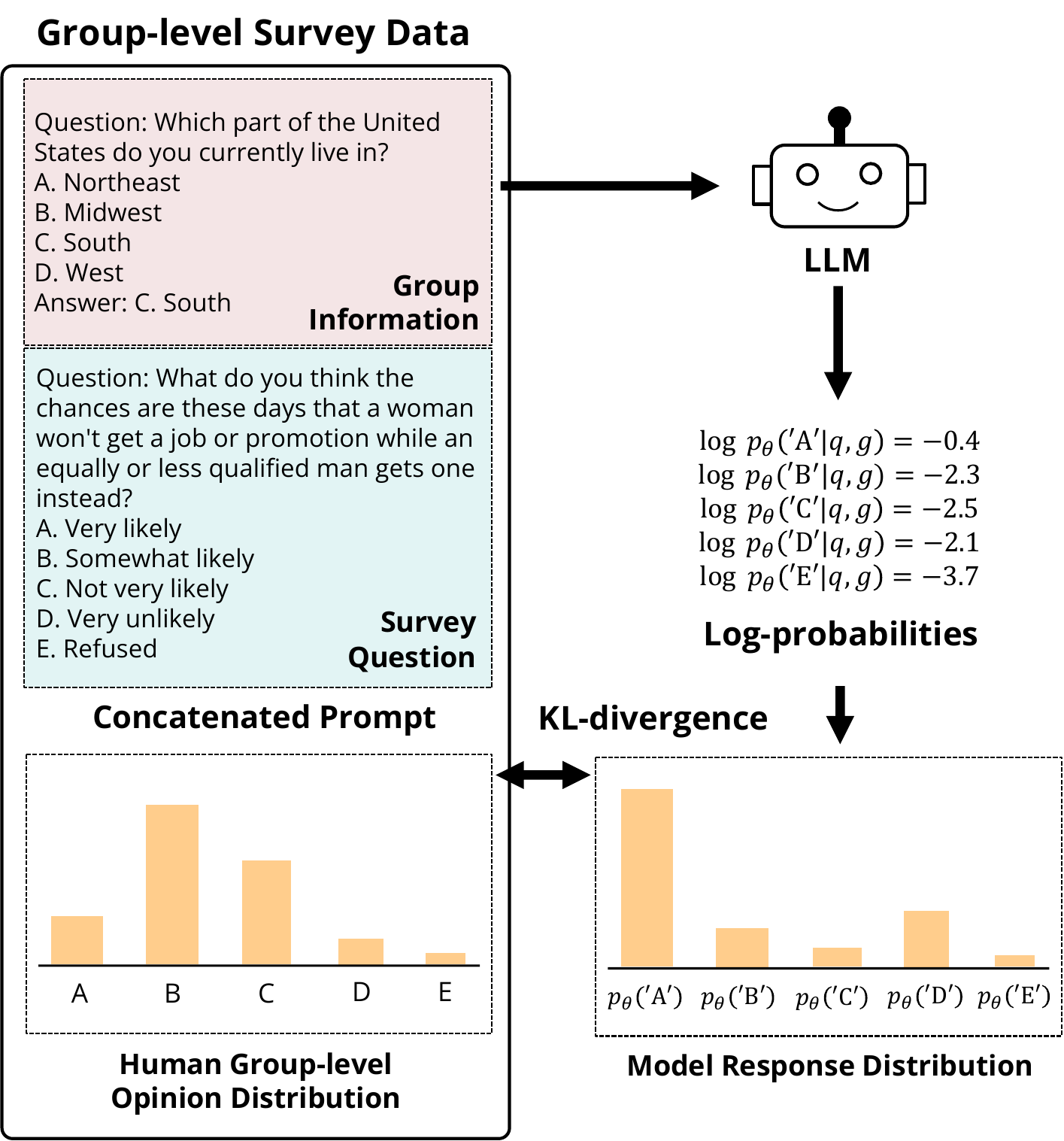}
    \caption{
    Proposed supervised fine-tuning setup with a survey response dataset such as \OURDATA.
    Survey data is 3-tuple of a survey question, target subpopulation information, and the observed human opinion distribution (\textit{i.e.} how subjects in the group responded to the given question).
    The training objective, $\mathcal{L}(\theta)$, is a forward KL divergence loss on language model predicted distribution of question option likelihoods; our loss guides the model predictions to match the response distribution of the specified human subpopulation.
    }
    \label{fig:method}
\end{figure}

\section{Methods}
\label{section_method}

\subsection{Fine-tuning LLMs on Human Response Distributions}
\label{section_method_fine_tuning_objective}
Our goal is to fine-tune an LLM to predict the distribution of responses for a multiple-choice question, conditioned on descriptions of a human subpopulation we want to simulate, typically a specific demographic, socioeconomic, or ideological subgroup. 
Consider the example in \Cref{fig:method}: the question asks, “What do you think the chances are these days that a woman won't get a job or promotion while an equally or less qualified man gets one instead?” The available responses are: \emph{A. Very likely, B. Somewhat likely, C. Not very likely, D. Very unlikely, and E. Refused}. 
In this case, the LLM will output a probability for each of the tokens corresponding to the choices A through E, thereby generating a complete response distribution that we aim to align with the true distribution observed in survey data.

Formally, let \(q \in Q\) be a question, \(g \in G\) be a subpopulation, and \(\mathcal{A}_q\) be the set of possible choices for question \(q\). 
An LLM with parameters \(\theta\) produces a conditional probability distribution \(p_{\theta}(\mathcal{A}_q \mid q, g)\). 
We fine-tune this model so that its predicted distribution for each \((q, g)\) mirrors the human response distribution \(p_H(\mathcal{A}_q \mid q, g)\) collected from real survey data.
To accomplish this, we apply LoRA fine-tuning~\cite{hu2021lora} and use the forward Kullback--Leibler (KL) divergence as our loss. 
Concretely, if \(p_H(\mathcal{A}_q \mid q, g)\)
represents the group-level empirical distribution of human opinions and \(p_{\theta}(\mathcal{A}_q \mid q, g)\) represents the model’s predicted distribution, our training objective is:
\[
\textstyle
\mathcal{L}(\theta) 
= \mathbb{E}_{q,g} \Bigl[
  D_{\mathrm{KL}}\bigl(
    p_H(\mathcal{A}_q \mid q, g) \,\big\|\, p_{\theta}(\mathcal{A}_q \mid q, g)
  \bigr)
\Bigr],
\]
where \(D_{\mathrm{KL}}\) denotes the KL divergence. In the example shown in \Cref{fig:method}, the model is trained to reduce the KL divergence between the target (survey-based) distribution over \{\(A, B, C, D, E\)\} and its predicted distribution for the subpopulation living in the Southern United States.

We choose forward KL (i.e., \(\mathrm{KL}\bigl(p_H\mid\mid p_\theta\bigr)\)) since it is sensitive to cases where \(p_H\) assigns high probability but \(p_\theta\) does not, naturally encouraging the model to \emph{cover} the real distribution.
This property aligns with standard maximum-likelihood training, where the model is penalized for underestimating any response that is frequent in the data.
In other words, if many participants in group \(g\) choose option ``A'' for question \(q\), then the model probability on ``A'' should be correspondingly high.

Instead of explicitly modeling the group response distribution as \(p_H(\mathcal{A}_q | q, g)\),
one could do two alternatives.
(1) One-hot encoding:
this approach \cite{li2024culturellm} approximates the distribution by a one-hot vector,
assigning a value of one to the most probable option and zero elsewhere.
(2) Data augmentation by response frequency:
this approach \cite{zhao2023group} expands the dataset by replicating
question-choice pairs in proportion to their observed frequency.
We adopt the explicit distribution modeling in our main experiments because it directly encodes the distributional information without requiring discrete sampling or replicating data points.
A detailed comparison of these approaches is provided in Section~\ref{section_experiments_modeling_response_distribution}.

\subsection{\OURDATA: a Comprehensive Survey Dataset to Fine-tune and Evaluate LLMs}
\label{section_method_datasets}
OpinionQA~\cite{santurkar2023whose} is a widely used dataset for fine-tuning and evaluating large language models (LLMs) on opinion prediction, containing roughly 500 questions drawn from 14 American Trends Panel (ATP) waves~\cite{atp}. 
Although valuable, it faces two important limitations:
(1) Limited thematic diversity—for instance, wave 26 focuses on the topic of firearms.
(2) Reliance on a single survey family (ATP), which risks overfitting to a particular style of questions and limits out-of-distribution evaluation on other sources (e.g., GSS).

To address these limitations, we introduce a new dataset, \OURDATA, that broadens both the thematic and institutional scope of opinion prediction data. 
For training, \OURDATA comprises 3,229 multiple-choice questions drawn from ATP waves 61--132, excluding 
waves included in OpinionQA.
In Table \ref{table:subpop-train-detail}, we list the topics of the ATP waves in \OURDATA vs. OpinionQA, both showing the increased thematic diversity of \OURDATA (with over 20 new topics) and the remaining unseen topics in OpinionQA that allow us to test whether LLMs fine-tuned on \OURDATA can generalize to unseen topics.

For evaluation, \OURDATA also includes 133 multiple-choice questions from the General Social Survey (GSS)~\cite{davern2024gss}, serving as an out-of-distribution benchmark.
This expanded collection not only broadens the range of topics beyond OpinionQA’s initial 500 questions, but also enables evaluation on surveys created and administered by different institutions (Pew Research Center vs. NORC-Chicago). 
Dataset curation and refinement pipeline is available in Appendix~\ref{appendix_dataset}.

\subsection{Evaluation Metric}
\label{section_method_evaluation_metric}
We use Wasserstein distance (WD) to quantify how closely the model’s predicted opinion distribution matches human survey data~\cite{santurkar2023whose, moon-etal-2024-virtual, meister2024benchmarking, zhao2023group}.
Formally, for a group $g$ representing some subpopulation and a question $q$
WD is defined as $\mathcal{WD}_{\theta}(q,g) = \mathcal{WD}(p_H(\mathcal{A}_q|q,g) , p_{\theta}(\mathcal{A}_q|q,g))$ (see formula in \Cref{appendix_training_train_objective}).
Since WD is computed over ordinal values, we map the categorical answer options to numbers, such as mapping ``Very likely'' to 1, ``Likely'' to 2, and so on. 

Some prior work utilizes one-hot accuracy~\cite{feng-etal-2024-modular, li2023steerability} as an evaluation metric.
However, 
one-hot accuracy only verifies whether the top-predicted choice matches the top human response,
thereby discarding distributional information.
In contrast, WD accounts for partial overlaps among the categories and reflects the ‘cost’ of shifting probability mass, providing a more nuanced assessment of distribution discrepancy.
Consider the example question provided in \Cref{fig:method}, where the human response distribution indicates that option B (“Somewhat likely”) is the most probable. 
Now consider two cases in which the model incorrectly predicts the top choice. 
In the first case, the model assigns a high probability to option A (``Very likely''), while in the second case, it assigns a high probability to option D (``Very unlikely''). 
Although one-hot accuracy would treat both predictions equally as errors, WD differentiates between them by accounting for the ordinal relationship among the options, penalizing the second prediction more heavily for its larger deviation from the true distribution.

%% file: sections/s4_experiment.tex
\section{Experiments}
\label{section_experiments}
\input{tables/main_results}

\subsection{Bounds of WD and Baselines}
\label{section_experiments_baseline}
In this section, we describe the lower/upper bounds and two baseline methods against which we compare our method.

\paragraph{Lower and upper bounds.} We use a uniform distribution over all available choices to establish an upper bound of the WD between a predicted and the target response distribution.
To compute a lower bound, we sample a group of human respondents from the original human respondents to calculate the WD between the two, and perform bootstrapping to obtain a robust estimate.
This lower bound captures the intrinsic variance arising from the respondent sampling process in opinion surveys.

\paragraph{Baselines.} We compare our approach with two baseline methods: prompting and Modular Pluralism~\cite{feng-etal-2024-modular}. 
For prompting, we consider both zero-shot and few-shot methods. In zero-shot prompting, we steer the LLM using demographic prompt formats. 
Specifically, we employ three different formats following \citet{santurkar2023whose}: \texttt{QA}, \texttt{BIO}, and \texttt{PORTRAY}. 
For instance, to condition the LLM to a person living in the South of the US, the \texttt{QA} format uses a question-answer format as illustrated in \Cref{fig:method}; 
the \texttt{BIO} format conditions the model with a first-person narrative such as ``I currently reside in the South."; 
and the \texttt{PORTRAY} format uses a third-person narrative like ``Answer the following question as if you currently reside in the South.".

Few-shot prompting augments the prompt with a few examples of question-response distribution pairs alongside the demographic label~\cite{hwang2023aligning}. 
In particular, we select the top five few-shot examples from the \OURDATA training set based on cosine similarity computed by the embedding model. 
In our experiments, we represent the response distribution in JSON format and require the model to output its prediction in the same JSON format, following the approach in~\citet{meister2024benchmarking}.

Modular pluralism~\cite{feng-etal-2024-modular} fine-tunes multiple LLMs on distinct datasets to capture the viewpoints of different communities~\cite{feng2023pretraining}. 
For a given question, each fine-tuned LLM generates an opinion that reflects the perspective of the community it represents, and a separate black-box LLM aggregates these outputs to produce the final distributional response. 
Detailed implementation of the lower/upper bounds and the baselines is provided in~\Cref{appendix_baseline_detail}.

\subsection{Generalization to Unseen Topics and Survey Families}
\label{section_experiments_prediction_of_opinion_distributions}
In this section, we assess the ability of our fine-tuned LLMs to generalize to unseen data—both in terms of new topics and entirely different survey families.
To evaluate these aspects, we use OpinionQA to measure generalization to unseen topics, and \OURDATA-Eval to test generalization to a different survey family.
We fine-tune four LLMs (Llama-2-7B, Llama-2-13B, Mistral-7B, and Llama-3-70B) on \OURDATA-Train. We opt for pretrained LLMs rather than instruction-following models, as previous work has shown that pretrained models perform better on this task~\cite{moon-etal-2024-virtual}. A detailed comparison between these model types is provided in \Cref{appendix_chat}.

\paragraph{Summary of results.} 
\Cref{table:main_results} reports the average WD metrics computed over all demographic groups and survey questions, comparing our fine-tuned models against various baseline approaches.
Our experiments show that fine-tuning on \OURDATA-Train significantly outperforms all other methods, yielding a 32–46\% reduction in WD on OpinionQA and a 39–42\% reduction on \OURDATA-Eval compared to the best baselines. 
Notably, \OURDATA-Train is based on ATP data, while \OURDATA-Eval is derived from GSS surveys—two distinct survey families that can differ in respondent pools, calibration techniques, and other methodological factors, leading to non-trivial distribution shifts despite both being representative of the US population. 
Furthermore, our fine-grained analyses at the wave level (see \Cref{appendix_finegrain}) confirm that these trends persist even at more detailed levels of evaluation.

\paragraph{Comparison to zero- and few-shot prompting.}
We first compare the performance of prompting methods with our approach. Zero-shot prompting results in only modest WD improvements over the upper bound, with the largest gain observed for Llama-3-70B and negligible improvements for Llama-2-7B. 
Even when using few-shot prompting---where five example question-response distribution pairs are provided---the performance gains remain minimal. 
This may be partly due to an under-optimized prompt format (\textit{e.g.} requiring JSON output) and the inherent sensitivity of language models to prompt formatting~\cite{sclar2023quantifying, anagnostidis2024susceptible}. 
These findings underscore the need for methods, such as fine-tuning, that enable relatively reliable predictions of opinion distributions.

\paragraph{Comparison to Modular Pluralism.}
Modular Pluralism improves one-hot accuracy, reducing prediction error from 72.7\% (zero-shot prompting) to 55.6\% on OpinionQA, but underperforms in matching the full distribution of option choices, measured as WD.
This discrepancy in performance highlights the limitations of methods that train LLMs to identify only the most probable response rather than modeling the entire distribution of responses.
Opinions are inherently distributed: even within a particular subpopulation such as a single demographic subgroup, distribution of opinions cannot be captured as a single most likely response.
Moreover, instruction-tuned models that serve as a black-box LLM tend to assign high probabilities on only specific tokens~\cite{lin2022teaching, kadavath2022language, achiam2023gpt}, further pushing the generated distribution away from the human distribution.

\subsection{Generalization across Target Subpopulations}
\label{section_experiments_per_group}
\begin{figure*}[!t]
    \centering
    \captionsetup{font=small}
    \includegraphics[width=0.95\linewidth]{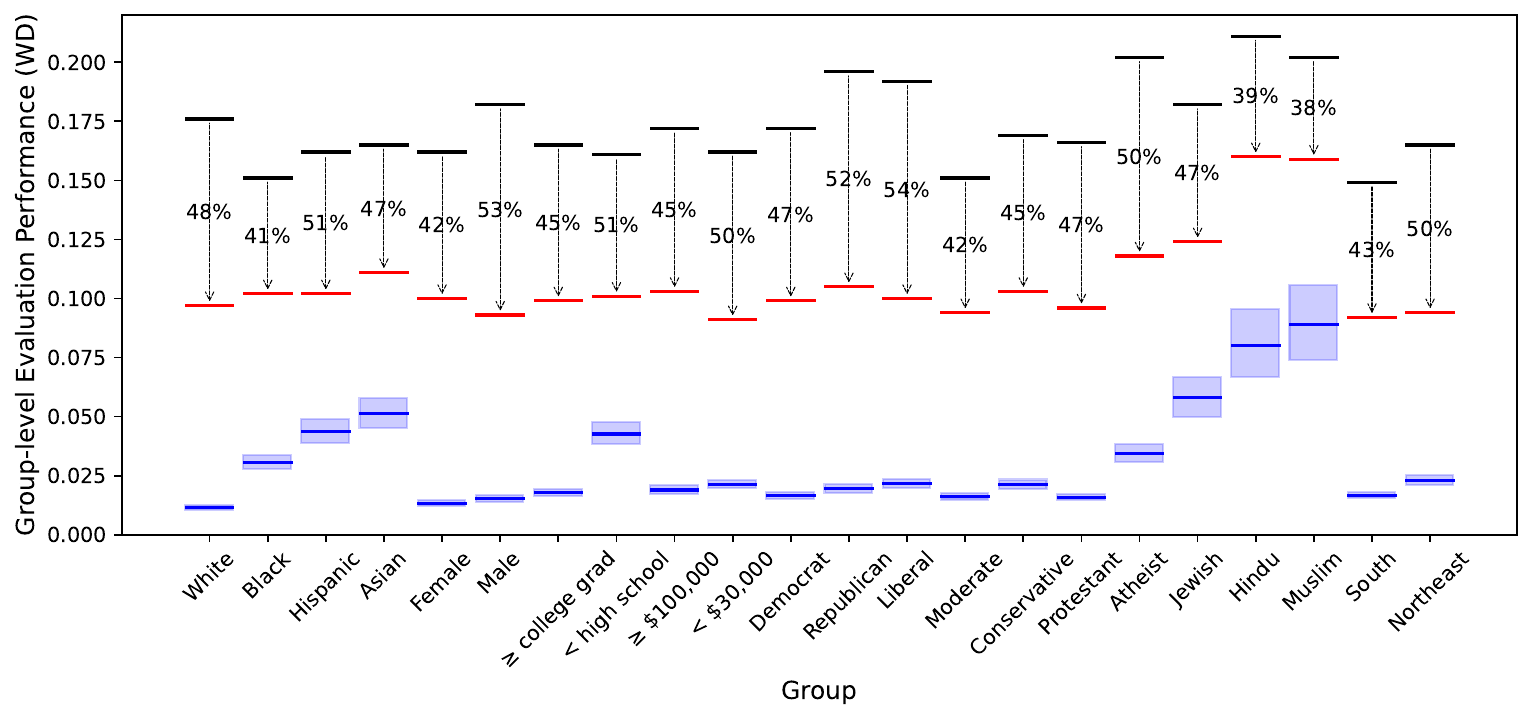}
    \caption{
    Per-group evaluation performance of our model Llama-2-7B-\OURDATA-FT (red lines) on OpinionQA.
    For comparison, the results from zero-shot QA prompting (black lines)
    and the lower bound (blue lines) are presented.
    We observe that the relative improvement,
    measuring how much of the gap between zero-shot prompting and the lower bound has been closed,
    remains consistent across subpopulations.
    Shaded blue regions represent the 95\% confidence interval of the lower-bound estimation for each group.
    Per-group results for other models (\Cref{table:per_group_opinionqa})
    and the results on \OURDATA evaluation set (\Cref{table:per_group_gss}) are available in \Cref{appendix_finegrain}.
    \label{fig:per_group_results}
    }
\end{figure*}

Here we report two key observations: 
(1) prediction performance improves consistently across most subpopulations represented in the fine-tuning data, and 
(2) the LLMs fine-tuned on \OURDATA-Train generalize well to subpopulations that were not included during fine-tuning.

\paragraph{Consistent performance improvements over subpopulations.}
\Cref{fig:per_group_results} shows the per-group WD on the OpinionQA evaluation for Llama-2-7B, 
comparing our fine-tuning approach with zero-shot prompting and the empirical WD lower bound. 
To evaluate the consistency of performance gains, we calculate the \textit{relative improvement} for each subpopulation as how much of the gap between zero-shot prompting and the empirical lower bound is reduced after fine-tuning. 
This measure allows us to account for varying lower bounds across subpopulations: since some groups have fewer respondents, there is greater uncertainty in their reported distribution in the survey data and greater variance between the original sample and bootstrap samples.

All 22 subpopulations demonstrate a large relative improvement after fine-tuning, ranging from 38\%--54\%.
The average relative improvement is 46.7\% with a standard deviation of 4.4\%.
This consistency confirms that our fine-tuning approach delivers balanced performance gains without disproportionately favoring any particular demographic subgroup.
We hypothesize that the consistent gains over groups largely stem from our dataset design, 
which allocates an equal number of training samples to each group. 
By ensuring uniformly distributed data points across subpopulations, the model captures sufficient subgroup-specific signals, ultimately leading to consistent performance improvements.

\paragraph{Generalization to unseen subpopulations.}
We further investigate how models fine-tuned with our approach and \OURDATA might show generalization to subpopulations that were not represented in the training data, a circumstance that may arise in real-world survey development. 
For the evaluation, we benchmark our methods against a zero-shot prompting baseline. 
Specifically, we evaluate our model, which is fine-tuned on 22 subpopulations provided in \OURDATA-Train,
on a set of subpopulations in OpinionQA that were not included in fine-tuning. 
This experiment not only checks generalization to unseen subpopulations, but also involves unseen survey questions, providing a robust assessment of the model capability for generalization to out-of-distribution data.

As shown in Table~\ref{table:unseen_demographics},
our model achieves a strong reduction in WD even for unseen subpopulations,
indicating that the model can be steered by demographic prompts beyond the seen subpopulations in training.
Interestingly, although \OURDATA-Train does not contain any data with opinion distributions of particular age groups
(\textit{e.g}. subjects of age 18-29 or those of age 65+),
the average relative improvement is 44.7\%, which is compatible with the average relative improvement for seen subpopulations. We provide results for other unseen groups in \Cref{table:unseen_demographic_all} of \Cref{appendix_unseen_subpopulation} (average relative improvement of 43.1\% with a standard deviation of 6.7\%).

\input{tables/unseen_demographics}

\paragraph{Steerability towards subpopulations.}
Given the large improvements in WD across subpopulations after fine-tuning, we want to test whether the LLM is truly adapting its predictions based on the subpopulation specified in its prompt (\textit{i.e.} the LLM is being steered) or if the improvements can be explained by the LLMs' predictions getting closer to human responses in general, without any subpopulation-specific adaptation. 
If the LLM is being steered, we should expect that the LLM's predictions for a target subpopulation $g_t$ are closer to the human distribution for $g_t$ when $g_t$ is the subpopulation specified in the prompt, compared to when another group $g_s$ is specified in the prompt.
We should also expect the gap in WD to be larger if the distance between the true human distributions for $g_t$ and $g_s$ are larger, such as differences between the youngest and oldest age groups compared to adjacent groups.

\begin{figure*}[!t]
    \centering
    \captionsetup{font=small}
    \includegraphics[width=1.00\linewidth]{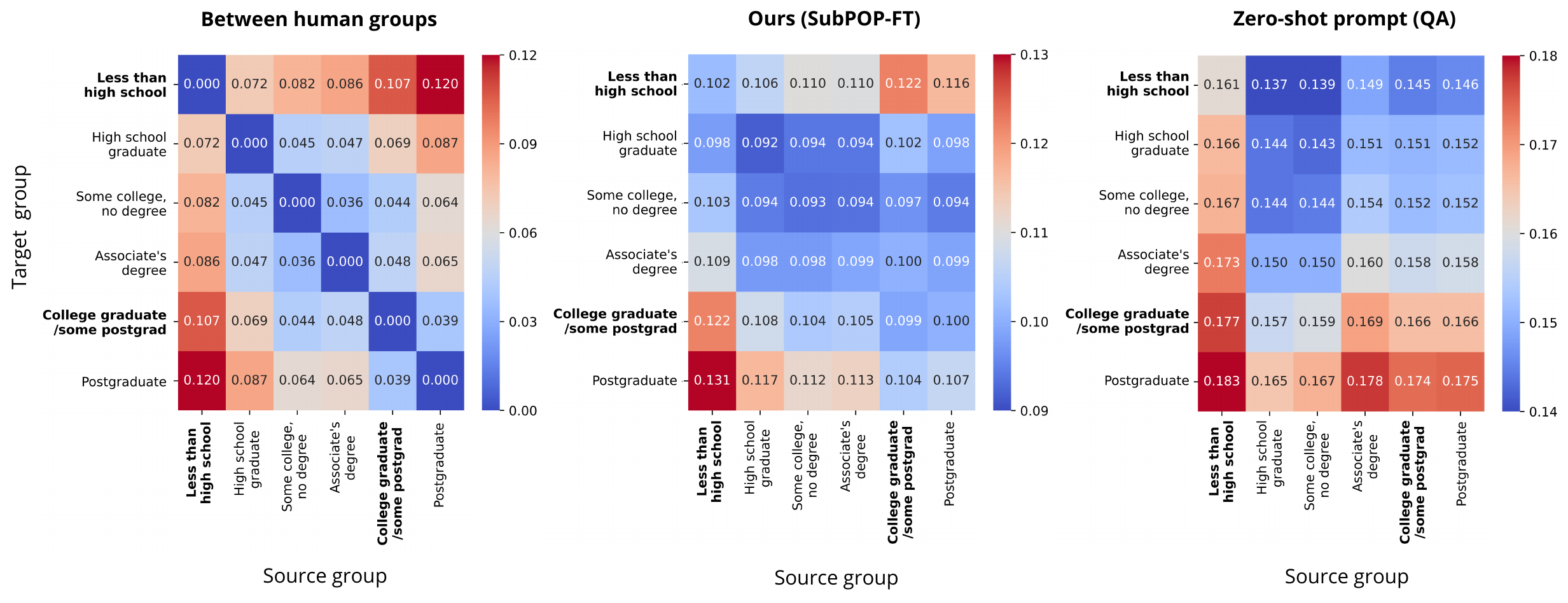}
    \caption{
    \textit{Intergroup disagreement} pattern between groups of different education levels calculated with OpinionQA and Llama-2-7B as a base model.
    A target human group is compared to
    (left) a source human group,
    (middle) our fine-tuned model conditioned on a source group,
    (right) a base model conditioned on a source group.
    Bold-faced groups are included in the fine-tuning data \OURDATA-Train, while the others aren't. In the human response (left), we observe a decreasing disagreement level as the education level becomes similar. This disagreement pattern exists in our fine-tuned model but not in the zero-shot prompting with a base model, indicating that our model can be steered to given subpopulation label even for unseen demographics while the base model cannot.
    }
    \label{fig:educ_heatmap}
\end{figure*}

Formally, we define the \textit{intergroup disagreement} between a target group $g_t$ and a source group $g_s$ as \(\mathcal{WD}(p_H(\mathcal{A}_q \mid q, g_t), p_H(\mathcal{A}_q \mid q, g_s))\) averaged over evaluation questions. In human responses (left of \Cref{fig:educ_heatmap}), the disagreement shows the pattern of locality: increases as the disparity in education levels between two groups grows.
We extend this notion to compare the human distribution from the target group \(g_t\) with the LLM-predicted distribution when the \textit{source} group $g_s$ is specified in the prompt,
\(\mathcal{WD}(p_H(\mathcal{A}_q \mid q, g_t), p_{\theta}(\mathcal{A}_q \mid q, g_s))\). If the model truly incorporates subpopulation information from the prompt, its intergroup disagreement pattern should mirror that of the human data.

Zero-shot prompting with the base model (right of \Cref{fig:educ_heatmap}) does not exhibit the locality pattern seen in the human data, indicating that it cannot be steered by subpopulation labels. In contrast, the fine-tuned model (middle of \Cref{fig:educ_heatmap}) reproduces a pattern resembling the human-human case, even though it was trained on only two education groups (“less than high school” and “college graduate/some postgrad”) and the other four groups were unseen. This result demonstrates that our fine-tuned model not only learns to condition on subpopulation information but also generalizes to subpopulations unseen during fine-tuning.
We provide the intergroup disagreement for other traits in \Cref{appendix_unseen_subpopulation}.

\subsection{Effect of Scaling the Dataset}
\label{section_experiments_scaling}
\begin{figure}[!t]
    \centering
    \captionsetup{font=small}
    \includegraphics[width=1.00\linewidth]{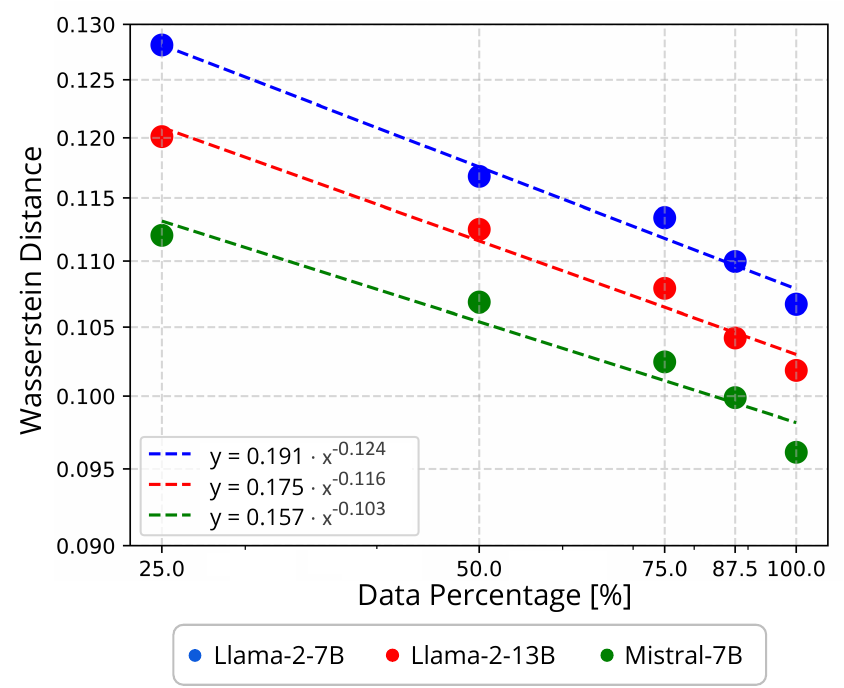}
    \caption{
    Evaluation results on OpinionQA after fine-tuning each LLM on increasingly large sampled subsets of \OURDATA-Train.
    Both axes are presented in a log scale.
    The $x$-axis is the size of sampled dataset and the $y$-axis is WD against human responses measured on OpinionQA.
    Dashed lines represent a line of best fit.
    Performances at data percentage of 100\% are identical to ours (\OURDATA-FT) in Table \ref{table:main_results}.
    }
    \label{fig:scaling}
\end{figure}

In this section, we examine performance scales with training dataset size. 
We randomly sample subsets containing 25\%, 50\%, 75\%, and 87.5\% of the full \OURDATA training set and evaluate three models—Llama-2-7B, Llama-2-13B, and Mistral-7B—on OpinionQA. 
As shown in \Cref{fig:scaling}, we observe diminishing marginal returns, as is typical with fine-tuning; for example, after training on a random 25\%, the models reach 72\%-78\% of the total improvement they achieve after fine-tuning on all of \OURDATA-train. However, the performance does not entirely plateau. Instead, it continues to improve as we further increase the training data from 25\% to 100\%. 
We fit linear trend lines (dotted in \Cref{fig:scaling}) to the results and observe that the slopes are similar for each model. 
This suggests that the rate of improvement—reflected by the slope in the power-law relationship—is intrinsic to the data and task rather than to the specific model architecture. 
In other words, LLMs exhibit comparable data efficiency, with performance gains that are fundamentally tied to dataset size rather than model-specific factors.

Using these trend lines, we can estimate the amount of fine-tuning data required to reach a target performance. For instance, we estimate that fine-tuning Mistral-7B on a dataset 25 times larger than the current \OURDATA training set would yield a WD value of 0.07, which is much closer to the empirical lower bound of 0.031 reported in \Cref{table:main_results}.
This result underscores the critical importance of collecting more high-quality data, as increased dataset size can drive significant improvements in model performance.

%% file: tables/main_results.tex
\begin{table*}[ht]
    \centering
    \scriptsize
    \caption{
    Evaluation on OpinionQA and the \OURDATA evaluation set (\OURDATA-Eval) for 22 subpopulations following ~\cite{santurkar2023whose}.
    We compute the WD by averaging over all questions and subpopulations.
    Lower and upper bounds of performance give guidance on how each method performs.
    For Modular Pluralism, we provide an error rate of one-hot prediction ($\dag$) (\Cref{section_method_evaluation_metric}) which was used in the original paper.
    }
    \label{table:main_results}
    \resizebox{1.0\textwidth}{!}{%
    \begin{tabular}{l|cccc|cccc}
    \toprule
     \multirow{2}{*}{Method} & \multicolumn{4}{c|}{\textbf{OpinionQA}}
     & \multicolumn{4}{c}{\textbf{\OURDATA-Eval}} \\
    & \textbf{Llama-2-7B} & \textbf{Llama-2-13B} & \textbf{Mistral-7B} & \textbf{Llama-3-70B}
    & \textbf{Llama-2-7B} & \textbf{Llama-2-13B} & \textbf{Mistral-7B} & \textbf{Llama-3-70B} \\
    \midrule
    Upper bound (Unif.) & \multicolumn{4}{c|}{0.178} & \multicolumn{4}{c}{0.208} \\
    Lower bound (Human) & \multicolumn{4}{c|}{0.031} & \multicolumn{4}{c}{0.033} \\
    \midrule
    Zero-shot prompt (\texttt{QA}) & 0.173 & 0.170 & 0.153 & 0.138 & 0.206 & 0.196 & 0.187 & 0.160 \\
    Zero-shot prompt (\texttt{BIO}) & 0.193 & 0.183 & 0.162 & 0.143 & 0.221 & 0.212 & 0.202 & 0.175 \\
    Zero-shot prompt (\texttt{PORTRAY})  & 0.195 & 0.207 & 0.158 & 0.209 & 0.212 & 0.242 & 0.194 & 0.247 \\
    Few-shot prompt  & 0.186 & 0.175 & 0.174 & 0.166 & 0.217   & 0.194   & 0.175 & 0.182   \\
    Modular Pluralism & \multicolumn{4}{c|}{0.285 ($^\dag$55.6\%)} & \multicolumn{4}{c}{0.279 ($^\dag$55.2\%)}   \\
    \highlightrow Ours (\OURDATA-FT) &  0.106 &  0.102 &  0.096 &  0.094 &  0.121 & 0.113 &  0.115 &  0.096 \\
    \bottomrule
    \end{tabular}
    }
\vspace{5pt}
\end{table*}

%% file: tables/unseen_demographics.tex
\begin{table}[t]
    \centering
    \scriptsize
    \captionsetup{font=small}
    \caption{
    Per-group evaluation performance of Llama-2-7B-\OURDATA-FT (Ours) on OpinionQA. 
    We report the lower bound, WD for zero-shot prompting, WD for Llama-2-7B-SubPOP-FT, and the relative improvement. 
    Rows highlighted in blue represent subpopulations included during fine-tuning, 
    while uncolored rows correspond to subpopulations that were unseen during fine-tuning.
    }
    \label{table:unseen_demographics}
    \resizebox{0.99\linewidth}{!}{
    \begin{tabular}{l|ccc|c}
    \toprule
    \multirow{2}{*}{\textbf{Group}} & Lower & Zero & \multirow{2}{*}{Ours} & Relative \\
    & Bound & Shot & & Improvement (\%) \\
    \midrule
    Age: 18-29 & 0.023 & 0.185 & 0.096 & 54.9 \\
    Age: 30-49 & 0.014 & 0.151 & 0.093 & 42.3 \\
    Age: 50-64 & 0.014 & 0.154 & 0.101 & 37.9 \\
    Age: 65+ & 0.013 & 0.195 & 0.115 & 44.0 \\
    \midrule
    \highlightrowtwo Less than high school & 0.043 & 0.161 & 0.101 & 50.8 \\
    High school graduate & 0.017 & 0.144 & 0.092 & 40.9 \\
    Some college, no degree & 0.018 & 0.144 & 0.093 & 40.5 \\
    Associate's degree & 0.026 & 0.159 & 0.098 & 44.9 \\
    \highlightrowtwo College grad & 0.018 & 0.165 & 0.099 & 44.9 \\
    Postgraduate & 0.015 & 0.174 & 0.106 & 42.8 \\
    \midrule
    Very conservative & 0.026 & 0.208 & 0.107 & 55.5 \\
    \highlightrowtwo Conservative & 0.021 & 0.169 & 0.103 & 44.6 \\
    \highlightrowtwo Moderate & 0.016 & 0.151 & 0.094 & 42.2 \\
    \highlightrowtwo Liberal & 0.022 & 0.192 & 0.100 & 54.1 \\
    Very liberal & 0.025 & 0.202 & 0.111 & 51.4 \\
    \midrule
    \highlightrowtwo Democrat   & 0.016 & 0.172 & 0.099 & 47.1 \\
    \highlightrowtwo Republican & 0.019 & 0.196 & 0.105 & 52.0 \\
    Independent & 0.016 & 0.155 & 0.093 & 44.5 \\
    Something Else & 0.026 & 0.162 & 0.092 & 51.0 \\
    \bottomrule    
    \end{tabular}
    }
\vspace{-5pt}
\end{table}

%% file: sections/s5_conclusion.tex
\section{Conclusion}
\vspace{-10pt}
In this work, we demonstrated that fine-tuning large language models on structured public opinion survey data markedly improves their ability to predict human response distributions. 
We curate \OURDATA—a dataset 6.5× larger than previous collections to fine-tune and evaluate LLMs on survey response distribution prediction task.
By fine-tuning on \OURDATA, we showed that LLMs can capture the nuanced, group-specific variability in public opinions, while also generalizing to unseen subpopulations, survey waves and question topics, and different survey families. 
Fine-tuning achieves consistent improvements across subpopulations of varying sizes, and our experiments demonstrate that fine-tuned LLMs are indeed \textit{adapting} their responses to the subpopulation specified in the prompt, even for subpopulations unseen during fine-tuning. 
Finally, our experiments also reveal that as the fine-tuning dataset grows, model performance continues to scale favorably, underscoring the value of our larger dataset. 

Generalization is a critical capability for LLMs, if they are to be used to assist public opinion research, as researchers are most in need of accurate opinion predictions for questions or subpopulations whom they have not surveyed before.
Our work, by greatly improving LLMs' ability to accurately predict opinions with fine-tuning and demonstrating strong generalization to out-of-distribution data, moves us closer towards the goal of leveraging LLMs for opinion prediction.
However, many open questions remain: 
why is the model able to generalize well to unseen subpopulations and questions, and under what conditions might it fail to do so?
How do we ensure that LLMs faithfully capture opinions along other dimensions that are not explored in this work, such as intersections of demographic identities or changing opinions over time? 
How should LLMs be integrated into survey designs, to serve as tools that can complement large-scale surveys with human participants? 
Answering these questions will require interdisciplinary collaborations with domain experts and critical assessments of LLMs' and traditional survey methods' strengths and weaknesses, so that we can most effectively and responsibly combine them to better estimate public opinions and inform public policies.

%% file: sections/a1_dataset.tex
\section{Dataset Details}
\label{appendix_dataset}

\subsection{American Trends Panel Datasets}
\label{appendix_atp_dataset}
Pew Research holds regular American Trends Panel (ATP) survey (called waves)~\cite{atp} covering various topics (\textit{e.g.} veterans, political priorities, gender and leadership) and releases result at an individual level.
For each anonymized individual, the following information is released: unique identification number, demographic details, survey responses, and weight.
Weights~\cite{mercer2018weighting} are the output of post-survey calibration process that helps adjusting survey results for response bias (e.g., non-response bias, sampling bias) correction and population representativeness.
As of January 2025, survey data until wave 132 has been released. About 20 surveys are conducted in each year.

\subsection{OpinionQA}
\label{appendix_opinionqa}
OpinionQA is a subset of ATP curated in \cite{santurkar2023whose}. This dataset consists of contentious 500 questions sampled from 14 ATP waves which have high intergroup disagreement (i.e. large Wasserstein distances among subpopulations' responses to a question). It also comes with hand-crafted ordinality information which provides structure to option lists. For example, options `Major reason', `Minor reason', and `Not a reason',
are assigned an ordinality mapping to 1, 2, and 3, respectively.
This ordinality allows a calculation of 1-dimensional Wasserstein distance.

Subpopulations we employ are listed in \Cref{table:opinionqa_detail}. This set of groups is adopted for several small-scale analysis \cite{santurkar2023whose, zhao2023group, kim2024few}.
We note that our approach is not limited to a specific number of groups and data is available for small or fine-grained demographic subpopulations.

\input{tables/opinionqa_detail}
\newpage
\input{tables/subpop-train-detail}

\subsection{\OURDATA-Train}
\label{appendix_ourdata_atp}
\begin{figure}[ht]
    \captionsetup{font=small}
    \includegraphics[width=\linewidth]{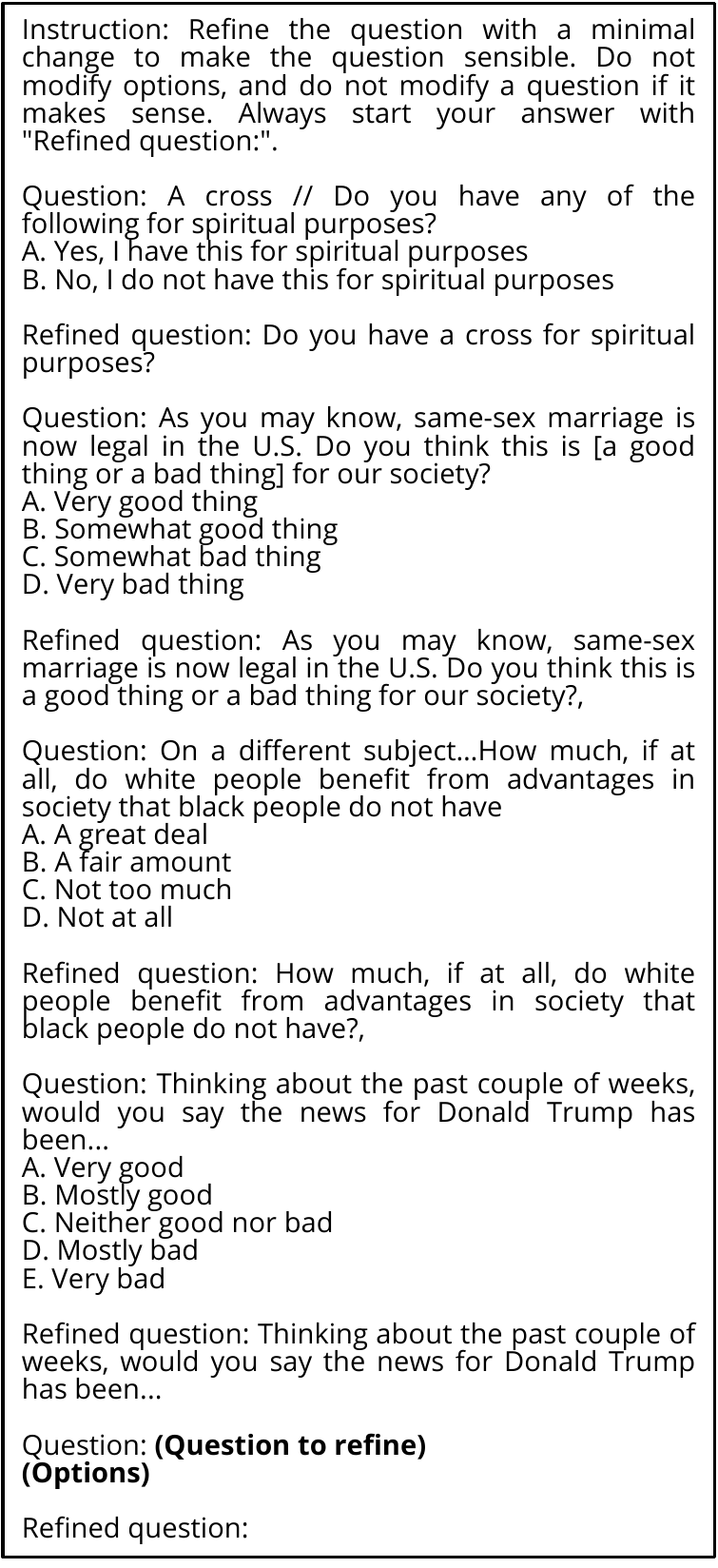}
    \vspace{-15pt}
    \caption{
    Few-shot prompt for refining the question to suit a language model prompting. An instruction is designed to make a minimal change to the original question, and in-context examples are provided.
    }
    \label{fig:question_refine_ice}
    \vspace{-5pt}
\end{figure}

We gather additional data from the American Trends Panel, specifically collecting 53 waves from Wave 61 to 132.
There are 62 waves from Wave 61 - 132, however, some waves have missing demographic or ideology information (for example, wave 63 does not contain political ideology information) or the data is not available hence removed during the curation process.
To refine the dataset, we exclude questions that meet the following criteria:
those with more than 10 response options, redacted response data, or dependencies on prior questions (e.g., assessing political strength). 
For the remaining questions, we use GPT-4o to refine their wording, ensuring they are well-suited for prompting the language models while making minimal modifications.
In \Cref{fig:question_refine_ice} we provide a few-shot prompt for question refinement.

\begin{figure}
    \centering
    \captionsetup{font=small}
    \includegraphics[width=0.8\linewidth]{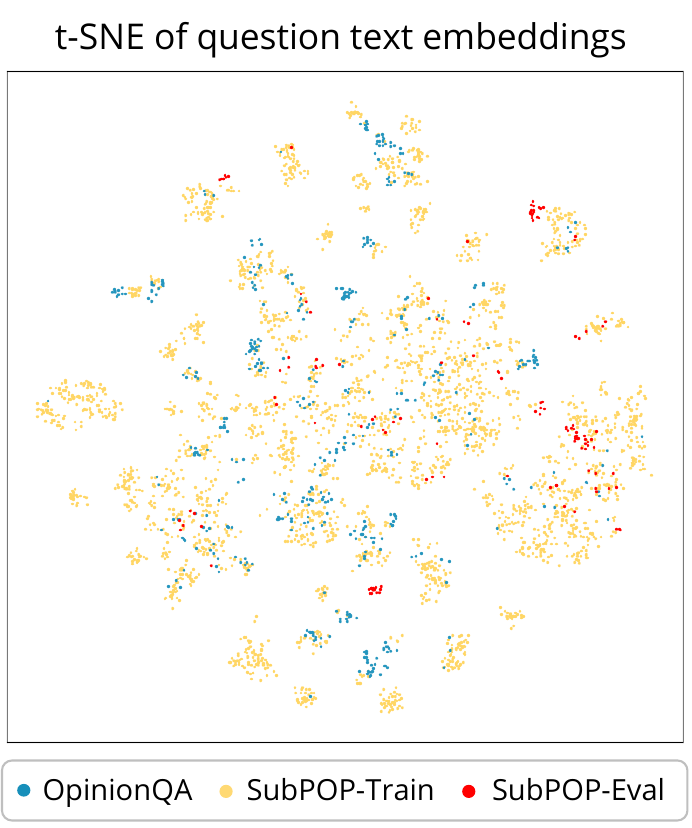}
    \caption{Embeddings of questions from OpinionQA, \OURDATA-Train, and \OURDATA-Eval.}
    \label{fig:question-embs}
\end{figure}

In Figure \ref{fig:question-embs}, we visualize the embeddings of the question texts (projected to 2-dimensions using t-SNE) from OpinionQA compared to \OURDATA-Train and \OURDATA-Eval.
The visualization shows how much larger our dataset is than OpinionQA (6.5$\times$), along with the expanded coverage of our dataset into semantic areas untouched by OpinionQA. 
The embeddings also reveal the distribution shift from ATP questions to GSS questions: while the ATP and GSS questions overlap in embedding space, the GSS question appear as small clusters, not evenly distributed over the ATP questions. 
In Table \ref{table:subpop-train-detail}, we list each ATP wave in \OURDATA-Train and OpinionQA, along with its number of questions and wave topic(s), as defined by ATP.\footnote{ATP wave topics and time periods are defined at \url{https://www.pewresearch.org/american-trends-panel-datasets/}.} 
The table indicates which topics are new in \OURDATA-Train compared to OpinionQA, indicating the expanded coverage of our dataset, along with which topics remain unseen in OpinionQA, which we can use to test LLMs fine-tuned on \OURDATA-Train for generalization.

\subsection{\OURDATA-Eval}
\label{appendix_outdata_gss}
To further evaluate the out-of-distribution generalization ability of our fine-tuned models, we subsample 133 questions from the GSS 2022 dataset \cite{davern2024gss}.
We apply the same selection criteria as outlined in ~\Cref{appendix_ourdata_atp}, excluding questions that are redacted, conditioned on prior questions, inferable directly from the group information, derived from a set of questions, or those with more than 10 response options.

\subsection{Inspection of Identical Questions}
\begin{figure}[!t]
    \captionsetup{font=small}
    \includegraphics[width=\linewidth]{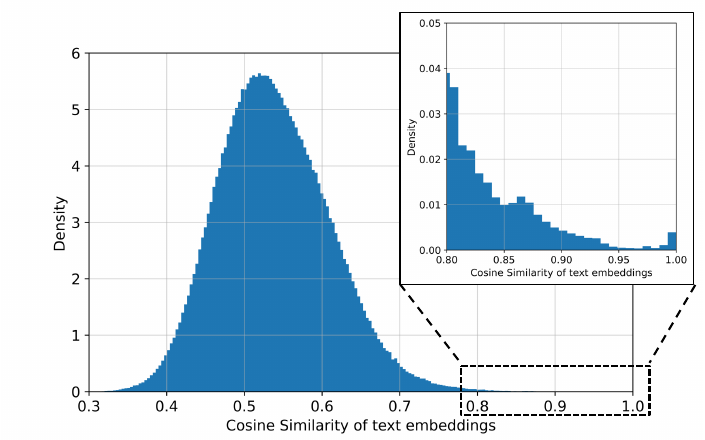}
    \vspace{-15pt}
    \caption{
    Distribution of cosine similarities between a question in \OURDATA-ATP and OpinionQA, having a long tail towards a high cosine similarity.
    We inspect the question pairs in the range of 0.8 to 1.0 (distribution shown in the magnified view) and use a similarity of 0.87 as a safe threshold to identify a semantically identical question pair.
    }
    \label{fig:cosine_sim_distribution}
    \vspace{-5pt}
\end{figure}

Distribution of cosine similarities between two text embeddings (an output of the embedding model OpenAI-text-embedding-3-large given a question text), one from a question in \OURDATA-Train and another from OpinionQA is shown in Figure~\ref{fig:cosine_sim_distribution}.
We observed a fraction of pairs having high cosine similarity,
and manually inspected question pairs with high relevance. We find that by setting a threshold cosine similarity of 0.87 we can detect all semantically identical pairs.
We took a conservative threshold of cosine similarity; this value was to maximize the recall at a cost of precision to ensure detection of overlapping questions.

%% file: tables/opinionqa_detail.tex
\begin{table}[ht]
    \centering
    \scriptsize
    \captionsetup{font=small}
    \caption{
    A list of 22 subpopulations used throughout our fine-tuning and analysis.
    We provide the number of respondents in each subpopulation in American Trends Panel Wave 82 for reference.
    }
    \vspace{-5pt}
    \label{table:opinionqa_detail}
    \begin{tabular}{ccc}
        \toprule
        \textbf{Trait} & \textbf{Groups} & \textbf{Population \% in Wave 82} \\
        \midrule
        \multirow{2}{*}{Region} & Northeast & 17.2 \\
        & South & 37.8 \\
        \midrule
        \multirow{2}{*}{Education} & College grad+ & 24.2 \\
        & Less than high school & 5.2 \\
        \midrule
        \multirow{2}{*}{Gender} & Male & 44.3 \\
        & Female & 54.6 \\
        \midrule
        \multirow{4}{*}{Race / ethnicity} & Black & 9.6 \\
        & White & 66.1 \\
        & Asian & 4.8 \\
        & Hispanic & 15.2 \\
        \midrule
        \multirow{2}{*}{Income} & \$100,000 or more & 21.8 \\
        & Less than \$30,000 & 21.3 \\
        \midrule
        \multirow{2}{*}{Political Party} & Democrat & 35.1 \\
        & Republican & 29.1 \\
        \midrule
        \multirow{3}{*}{Political Ideology} & Liberal & 20.0 \\
        & Conservative & 22.6 \\
        & Moderate & 38.3 \\
        \midrule
        \multirow{5}{*}{Religion} & Protestant & 40.8 \\
        & Jewish & 2.0 \\
        & Hindu & 0.9 \\
        & Atheist & 0.6 \\
        & Muslim & 0.7 \\
        \bottomrule    
        \end{tabular}
\vspace{-5pt}
\end{table}

%% file: tables/subpop-train-detail.tex
\begin{table}[H]
    \centering
    \scriptsize
    \captionsetup{font=small}
    \caption{
    American Trends Panel (ATP) wave topics for waves included in \OURDATA-Train (top) and OpinionQA (bottom).
    Golden rows represent wave topics in \OURDATA-Train that are not present in OpinionQA, and blue rows represent wave topics in OpinionQA that are not present in \OURDATA-Train.
    For waves 68-79, survey questions related to COVID-19 (\textit{e.g.}, contact tracing, vaccines, and relocation) were included as part of a survey along with the main survey topic.
    }
    \label{table:subpop-train-detail}
    \begin{tabular}{ccm{4.5cm}}
    \toprule
    \textbf{Wave} & \textbf{\# questions} & \textbf{Wave Topic} \\
    \midrule
    68 & 90 & American News Pathways, George Floyd, Black Lives Matter \\
    69 & 92 & Politics, 2020 Census \\
    \highlightrow 70 & 56 & Religion in public life, social media’s role in politics and society \\
    71 & 84 & Voter attitudes \\
    \highlightrow 72 & 18 & New media \\
    \highlightrow 73 & 82 & American News Pathways, social media \\
    74 & 51 & Online harassment, race relations \\
    75 & 18 & 2020 pre-election survey \\
    \highlightrow 76 & 44 & American News Pathways \\
    \highlightrow 77 & 13 & Culture of work \\
    78 & 57 & 2020 post-election survey \\
    \highlightrow 79 & 93 & American News Pathways \\
    80 & 45 & Political priorities \\
    \highlightrow 81 & 52 & Economics, pandemic financial outlook \\
    \highlightrow 83 & 54 & Coronavirus vaccines and restrictions \\
    \highlightrow 84 & 50 & Religion in politics and tolerance \\
    85 & 93 & News coverage of the Biden administration’s first 100 days \\
    87 & 90 & Current political news and topics \\
    \highlightrow 88 & 37 & Tech companies and policy issues \\
    \highlightrow 90 & 79 & Twitter news attitudes \\
    91 & 64 & Benchmark study \\
    \highlightrow 93 & 19 & Social media update \\
    95 & 78 & Politics timely and topical \\
    \highlightrow 96 & 57 & Post-coronavirus pandemic spirituality \\
    \highlightrow 98 & 76 & Coronavirus impacts on communities, living arrangements and life decisions \\
    \highlightrow 99 & 20 & Artificial intelligence (AI) and human enhancement \\
    \highlightrow 103 & 12 & Economic well-being \\
    104 & 92 & Politics, Religion in Public Life \\
    105 & 38 & Global Attitudes US Survey 2022 \\
    \highlightrow 106 & 62 & Religion and the environment \\
    107 & 92 & Government and Parties \\
    \highlightrow 108 & 83 & COVID and Climate, Energy and the Environment \\
    109 & 51 & New Digital Platforms and Gender Identity \\
    110 & 90 & Politics timely and topical \\
    \highlightrow 111 & 23 & Online dating and E-commerce \\
    \highlightrow 112 & 31 & Social media update \\
    113 & 53 & 2022 National Survey of Latinos (NSL) \\
    \highlightrow 114 & 93 & Covid, scientists, and religion \\
    115 & 63 & Parents survey \\
    116 & 75 & Politics timely and topical \\
    117 & 16 & Religion and politics \\
    118 & 25 & Podcasts, news, and racial identity \\
    \highlightrow 119 & 70 & AI and human enhancement \\
    120 & 61 & Politics timely and topical \\
    \highlightrow 121 & 31 & Culture of work \\
    124 & 75 & Global Attitudes US Survey 2023 \\
    125 & 69 & Politics timely and topical \\
    126 & 93 & Racial attitudes, modern family \\
    127 & 59 & Americans and their data \\
    \highlightrow 128 & 89 & Americans and planet Earth \\
    129 & 107 & Politics timely and topical \\
    130 & 94 & Politics representation \\
    131 & 70 & Gender and leadership \\
    \midrule    
    \textbf{Wave} & \textbf{\# questions} & \textbf{Wave Topic} \\
    \midrule
    \highlightrowtwo 26 & 44 & Guns \\
    29 & 20 & Views on gender \\
    32 & 24 & Community types, Sexual harassment \\
    \highlightrowtwo 34 & 16 & Biomedical and food issues \\
    36 & 68 & Gender and leadership \\
    \highlightrowtwo 41 & 41 & Views of America in 2050 \\
    \highlightrowtwo 42 & 26 & Trust in science \\
    43 & 51 & Race in America \\
    45 & 13 & Misinformation \\
    49 & 19 & Privacy and surveillance \\
    50 & 43 & American families \\
    54 & 50 & Economic inequality \\
    82 & 56 & 2021 Global Attitudes Project U.S. survey \\
    92 & 23 & Political Typology \\
    \bottomrule   
    \end{tabular}
\end{table}

%% file: sections/a2_sft.tex
\section{Experiment Details}
\label{appendix_training}
We conduct our experiments using Nvidia A100 GPUs with 80GB VRAM.
Hyperparameter tuning is performed over learning rates \{5e-5, 1e-4, 2e-4\}
and batch sizes \{64, 128, 256\}.
After evaluating possible combinations,
we select a (learning rate, batch size) = (2e-4, 256) for Llama-2-7B,
(learning rate, batch size) = (2e-4, 256) for Mistral-7B-v0.1,
and (learning rate, batch size) = (1e-4, 256) for Llama-2-13B
when utilizing the full training dataset.
For Llama-3-70B, we have not done hyperparameter search but heuristically used
(learning rate, batch size) = (2e-5, 256).
For sub-sampled training data (Figure \ref{fig:scaling}), we use the following configurations:
\begin{itemize}[leftmargin=*]
    \vspace{-7pt}
    \item (lr, bs) = (2e-4, 256) for 75\% of the training data
    \vspace{-7pt}
    \item (lr, bs) = (1e-4, 128) for 50\% of the training data
    \vspace{-7pt}
    \item (lr, bs) = (1e-4, 128) for 25\% of the training data
\end{itemize}

All training is performed using LoRA \cite{hu2021lora}, with LoRA parameters initialized from a normal distribution with $\sigma=0.02$. We set the LoRA rank to 8, alpha to 32, and apply a dropout rate of 0.05. LoRA weights are applied to the query and value matrices. The AdamW \cite{loshchilov2017decoupled} optimizer is used with a weight decay of 0.

We use offline batched inference of vLLM (version 0.7.2)~\cite{kwon2023efficient} for inference and measuring response probability distribution of all methods.

\paragraph{Choice of the Training Objective.}
\label{appendix_training_train_objective}
In this section, we explore both forward KL-divergence and Wasserstein Distance (WD) as training objectives. The forward KL-divergence is defined as
\[
D_{\mathrm{KL}}(p_H \| p_{\theta}) = \sum_{a \in \mathcal{A}_q} p_H(a) \log \frac{p_H(a)}{p_{\theta}(a)},
\]
where \(p_H(a) \equiv p_H(a \mid q, g)\) and \(p_{\theta}(a) \equiv p_{\theta}(a \mid q, g)\). Similarly, WD is given by
\[
\mathcal{WD}(p_H, p_{\theta}) = \min_{\gamma \in \Pi(p_H,p_{\theta})} \sum_{a,a' \in \mathcal{A}_q} \gamma(a,a')\, d(a,a'),
\]
with \(\Pi(p_H,p_{\theta})\) denoting the set of all couplings between \(p_H\) and \(p_{\theta}\), and \(d(a,a')\) the L1 distance between choices.
Since survey responses are inherently one-dimensional and ordinal, we can simplify the computation of WD using cumulative distribution functions (CDFs). In the 1-D case, WD is computed as
\begin{align*}
    \mathcal{WD}(p_H, p_{\theta}) 
    &= \int_{-\infty}^{+\infty} \left| F_{p_H}(x) - F_{p_{\theta}}(x) \right| \, dx, \\
    &= \sum_{i=1}^{n} \left| F_{p_H}(i) - F_{p_{\theta}}(i) \right|
\end{align*}
where \(F_{p_H}\) and \(F_{p_{\theta}}\) are the CDFs corresponding to \(p_H\) and \(p_{\theta}\), respectively.
We use this discrete formulation as the WD loss in our training.

\begin{figure}[!t]
    \captionsetup{font=small}
    \includegraphics[width=\linewidth]{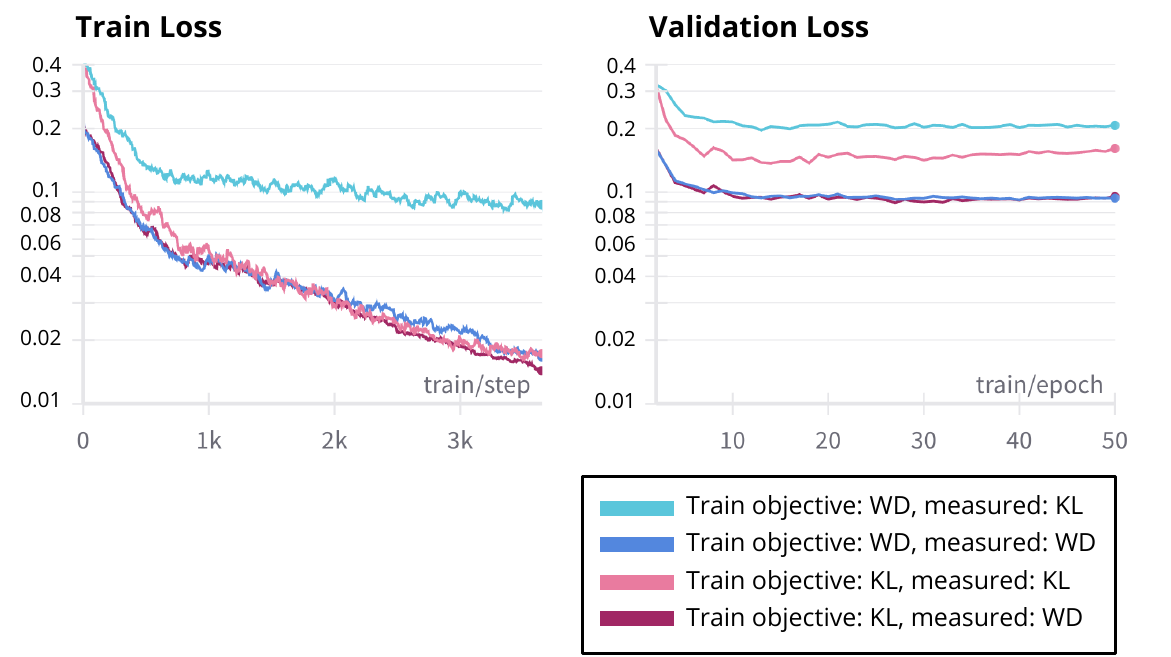}
    \caption{
    Train loss curve (left) and validation loss curve (right) for Llama-2-7B
    fine-tuned on 90\% of OpinionQA, with the remaining 10\% used for validation.
    Light and dark blue lines represent KL-divergence (KL) and Wasserstein distance (WD) when used KL as a training objective,
    while light and dark red lines represent KL and WD when used WD as a training objective.
    The two training objectives yield similar results in terms of WD, the primary measure of opinion distribution matching in our work.
    }
    \label{fig:wd_vs_ce_loss}
\end{figure}

While training with WD resulted in a higher KL-divergence on the validation set, the validation WD converged to similar levels regardless of the objective (see Figure \ref{fig:wd_vs_ce_loss}). We attribute this to KL-divergence penalizing low-probability assignments without significantly altering the overall distribution geometry. Given the KL divergence's broader applicability—without requiring ordinal information—we primarily used KL-divergence in our experiments.

%% file: sections/a3_chat.tex
\section{Additional Experiments}
\subsection{Effect of Response Distribution Modeling}
\label{section_experiments_modeling_response_distribution}

In this section, we compare different methods for capturing the distribution of human responses. We consider three approaches:
\begin{enumerate}
    \item \emph{One-hot}: Predicting only the most probable response, which ignores the full distribution over all responses~\cite{li2024culturellm}.
    \item \emph{Augment by N}: Augmenting the dataset by replicating each response by a factor of N according to its observed frequency~\cite{zhao2023group}.
    \item \emph{Explicit probability modeling}: Directly modeling the full response distribution using the actual probability values for each option.
\end{enumerate}

\Cref{table:response_distribution_modeling} summarizes the results of these approaches. Notably, explicit probability modeling substantially outperforms the one-hot method, demonstrating that simply predicting the single most frequent response fails to capture the opinion diversity present within each subpopulation.

Compared with augment by \(N\) (2nd and 3rd column in \Cref{table:response_distribution_modeling}), explicit probability modeling also achieves better performance. 
Importantly, the performance gap exceeds the quantization error introduced by discretizing the response distribution. 
For instance, when discretizing with a factor of \(N\), the quantization error is \(\frac{1}{2N}\)—approximately 0.01 or 0.005 in the cases shown in \Cref{table:response_distribution_modeling}. 
Moreover, explicit modeling offers the practical benefit of reducing the data volume by a factor of \(N\) compared to the augmentation approach, thereby lowering the computational cost of fine-tuning LLMs.

These results underscore the importance of explicit distribution modeling. 
By aligning the model’s predictive distribution directly with the survey distribution, we achieve higher accuracy with fewer data samples, avoiding the rounding errors and replication overheads that are inherent to data-augmentation approaches.

\input{tables/response_distribution_modeling}

\subsection{Post-trained Model}
\label{appendix_chat}
We fine-tune Llama-2-7B-chat to observe the effect of starting 
from checkpoints that have been instruction-tuned via Reinforcement Learning from Human Feedback (RLHF).
\Cref{table:chat_performance} shows the evaluation performance of a baseline method (Zero-shot prompt (\texttt{QA})), fine-tuned base model (Llama-2-7B) and fine-tuned chat model (Llama-2-7B-chat).
We observe the significant performance improvement,
while the baseline method performs worse then the models not instruction-tuned (\Cref{table:main_results}).
Especially, the performance for \OURDATA-Eval of chat model is significantly worse than that of base model.
We observe the high WD of the baseline method resulting from the model assigning high probability to a specific token (e.g. `A'),
being far apart from the human opinion distribution.
After fine-tuning the model are able to generate a more distributed probability of answer tokens.
This result coincides with the result reported in~\cite{moon-etal-2024-virtual}.

\input{tables/chat_performance}

\subsection{Generalization to Unseen Subpopulations}
\label{appendix_unseen_subpopulation}
Here we present a complete list of evaluation performance on OpinionQA
for unseen subpopulations (the groups not used to fine-tune our model)
and perform an analysis that shows
our fine-tuned models are able to steer towards the given subpopulation information.

\begin{figure*}[!t]
    \captionsetup{font=small}
    \includegraphics[width=\linewidth]{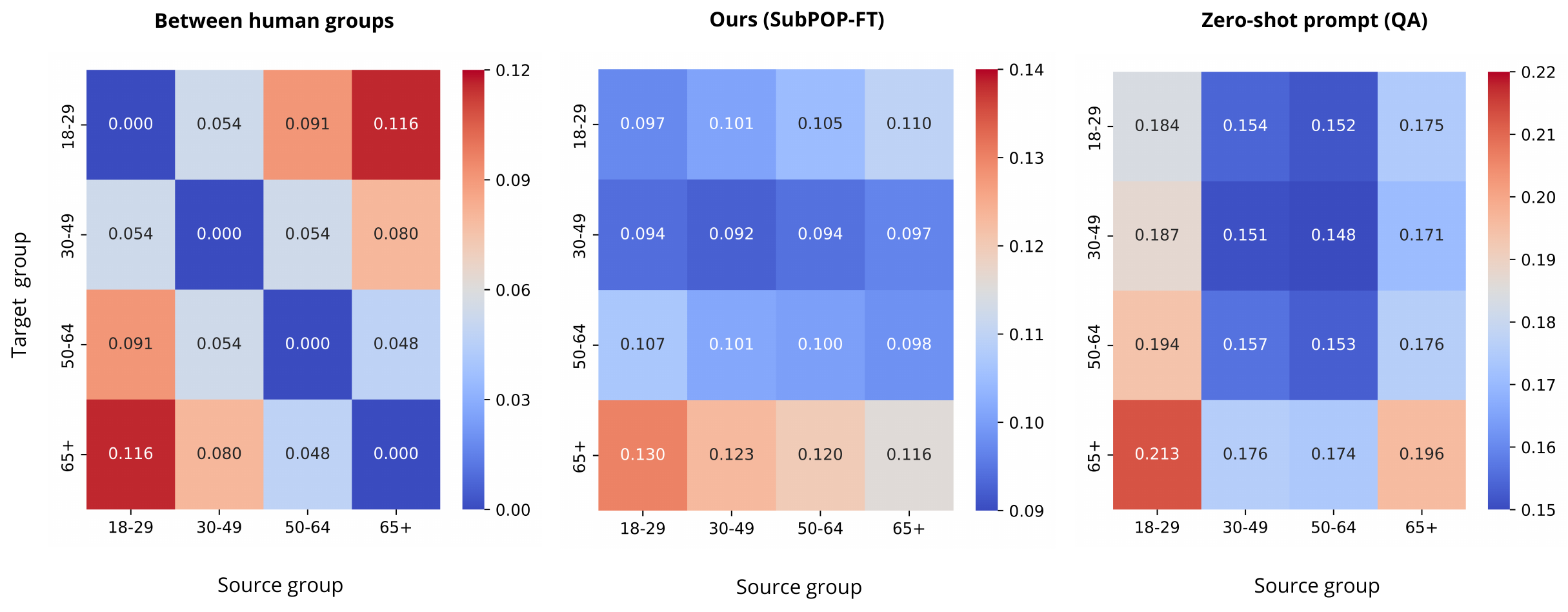}
    \caption{
    Heatmap of intergroup disagreement between a target human group ($y$-axis)
    and a source group ($x$-axis, either a human group or a group simulated with the language model),
    for OpinionQA evaluation data and age trait using Llama-2-7B as a base model.
    All subpopulations are unseen during fine-tuning.
    }
    \label{fig:age_heatmap}
\end{figure*}

\begin{figure*}[!t]
    \captionsetup{font=small}
    \includegraphics[width=\linewidth]{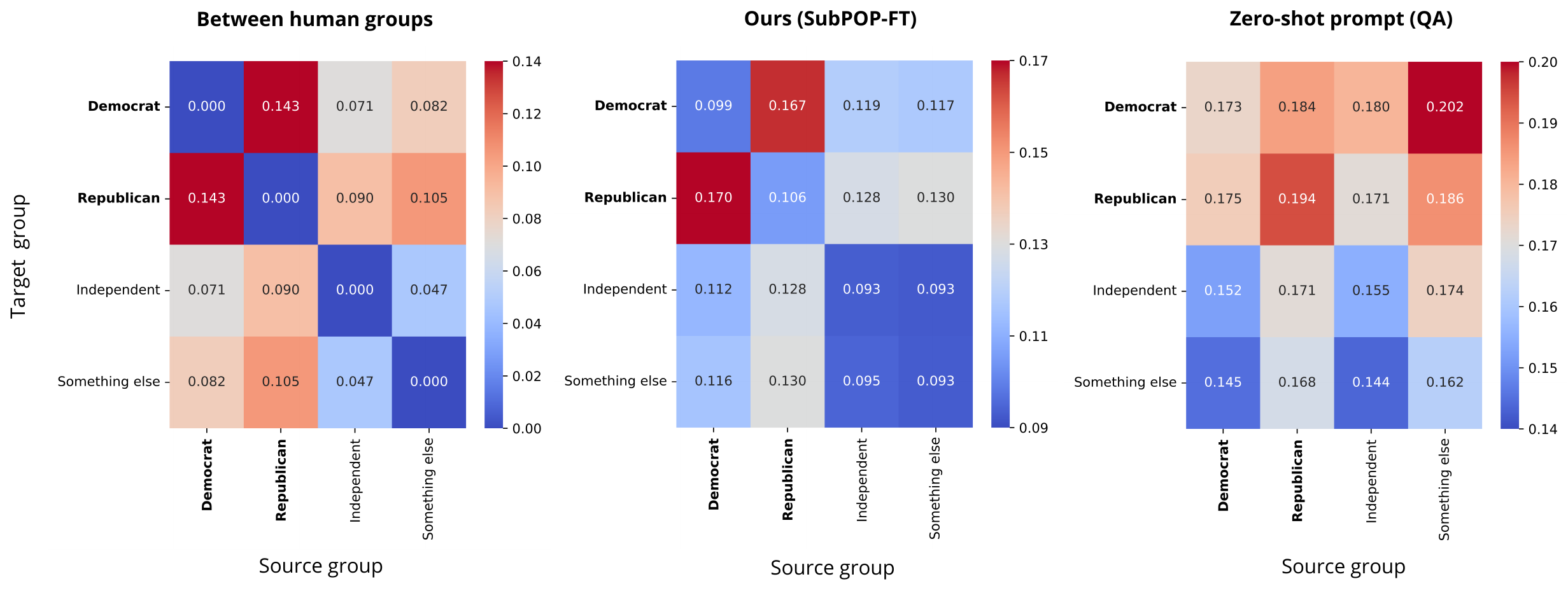}
    \caption{
    Heatmap of intergroup disagreement between a target human group ($y$-axis)
    and a source group ($x$-axis, either a human group or a group simulated with the language model),
    for OpinionQA evaluation data and political party (affiliation) trait using Llama-2-7B as a base model.
    Two subpopulations, Democrat and Republican, are seen during fine-tuning, while Independent and Something Else are unseen.
    \vspace{20pt}
    }
    \label{fig:polparty_heatmap}
\end{figure*}

\begin{figure*}[!t]
    \captionsetup{font=small}
    \includegraphics[width=\linewidth]{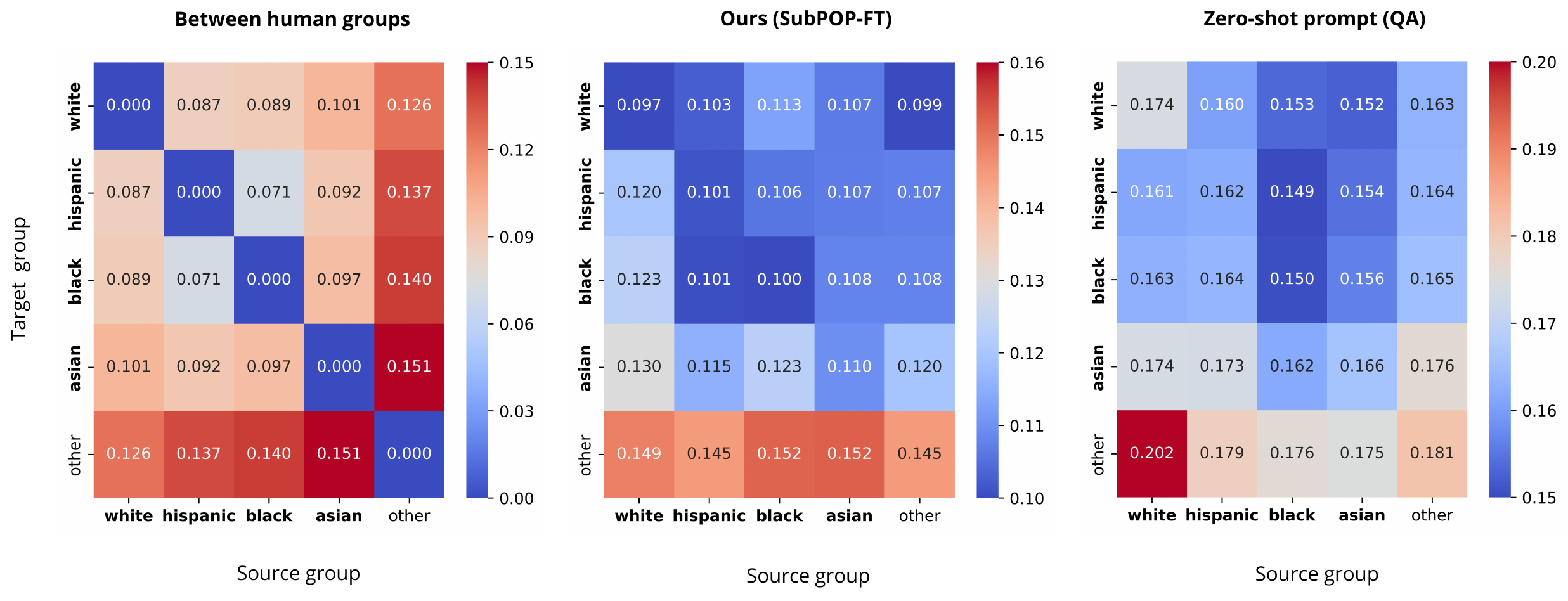}
    \caption{
    Heatmap of intergroup disagreement between a target human group ($y$-axis)
    and a source group ($x$-axis, either a human group or a group simulated with the language model),
    for OpinionQA evaluation data and race / ethnicity trait using Llama-2-7B as a base model.
    Four subpopulations except `Other' are seen during fine-tuning.
    In this case, the model does not well predict the opinions of Other group.
    We suspect this occurs because Other is a group with highly diverse race or ethnicity backgrounds,
    making it inherently difficult to infer its opinion distribution from those of White, Hispanic, Black, and Asian subpopulations.
    }
    \label{fig:race_heatmap}
\end{figure*}

\begin{figure*}[!t]
    \captionsetup{font=small}
    \includegraphics[width=\linewidth]{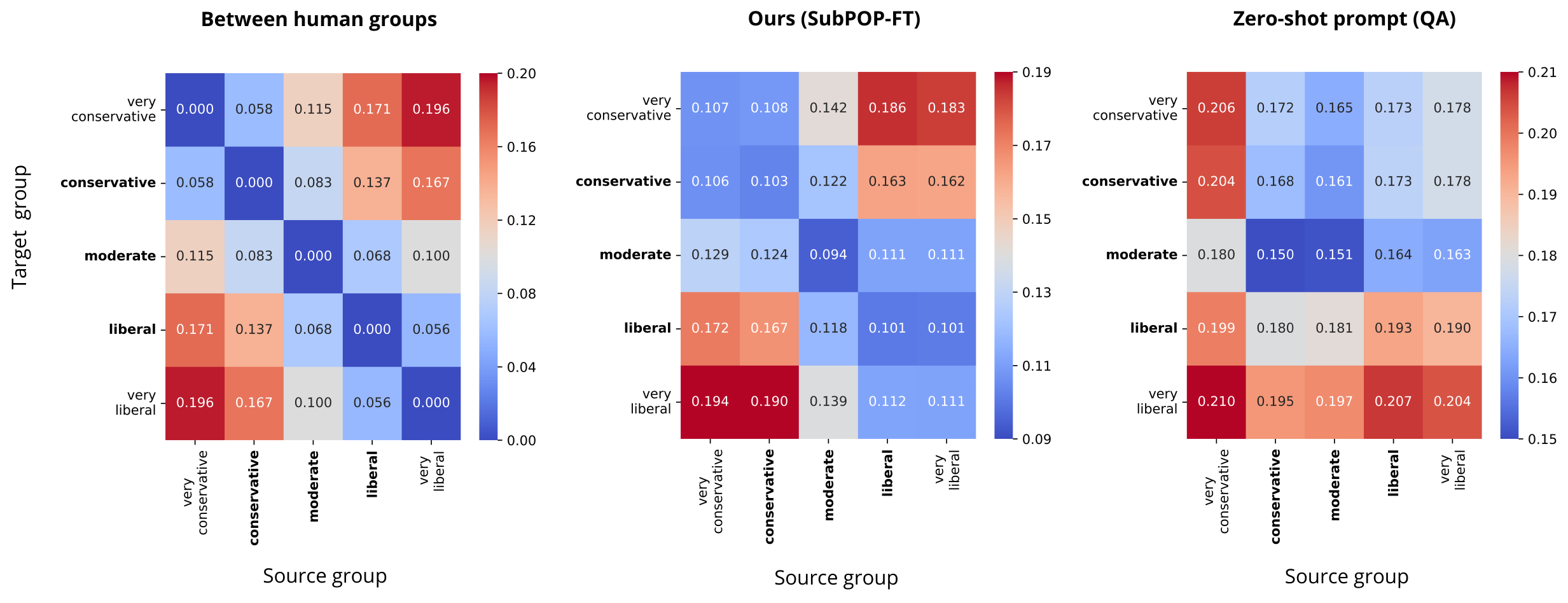}
    \caption{
    Heatmap of intergroup disagreement between a target human group ($y$-axis)
    and a source group ($x$-axis, either a human group or a group simulated with the language model),
    for OpinionQA evaluation data and political ideology trait using Llama-2-7B as a base model.
    Three subpopulations, Conservative, Moderate, and Liberal are seen during fine-tuning, while Very conservative and Very liberal are not seen.
    }
    \label{fig:polideology_heatmap}
\end{figure*}

As shown in \Cref{table:unseen_demographic_all}, we observe a performance improvement across unseen subpopulations.
To verify that the performance improvements arise from the fine-tuned model being able to steer towards given subpopulations, we measure \textit{intergroup disagreement} pattern for the demographic and ideology traits, shown in
\Cref{fig:age_heatmap},
\ref{fig:polparty_heatmap},
\ref{fig:race_heatmap},
and \ref{fig:polideology_heatmap}.
We consistently observe across traits that the disagreement pattern of our model resembles that of the human group, while zero-shot prompting with the base model exhibits a pattern completely different from the human group result.
This observation shows that our fine-tuned model learns to condition on subpopulation information and also generalizes to subpopulations unseen during fine-tuning.

\input{tables/unseen_demographics_all}


%% file: tables/response_distribution_modeling.tex
\begin{table}[b]
    \centering
    \scriptsize
    \captionsetup{font=small}
    \caption{
    Comparison of evaluation performance for three response distribution modeling approaches,
    with Llama-2-7B as a base model.
    The last column (Explicit) is identical to the ours presented in \Cref{table:main_results}.
    A model fine-tuned to predict the most probable choice (one-hot) performs the worst,
    as the model has not learned distributional opinion at fine-tuning phase.
    A model trained on augmented data (Aug. ($\times$50, $\times$100)),
    while performing much better than one-hot still underperforms the explicit distribution modeling.
    }
    \label{table:response_distribution_modeling}
    \resizebox{1.0\linewidth}{!}{
    \begin{tabular}{c | cccc}
    \toprule
    Eval Dataset & 
    One-hot & 
    Aug. ($\times$ 50)  &
    Aug. ($\times$ 100) &
    Explicit (Ours) \\
    \midrule
    OpinionQA
    & 0.163 & 0.110 & 0.107 & 0.106 \\
    \midrule
    \OURDATA-Eval
    & 0.178 & 0.130 & 0.123 & 0.121 \\
    \bottomrule
    \end{tabular}
    }
\end{table}

%% file: tables/chat_performance.tex
\begin{table}[ht]
    \centering
    \scriptsize
    \caption{
    Performance of the fine-tuned Llama-2-7B-chat model (Chat LLM FT).
    For comparison, we also present lower and upper bounds, the baseline method Zero-shot prompt (\texttt{QA}) and our fine-tuned Llama-2-7B (Base LLM FT).
    }
    \label{table:chat_performance}
    \begin{tabular}{l|c|c}
    \toprule
    Method & \textbf{OpinionQA} & \textbf{\OURDATA-Eval} \\
    \midrule
    Upper bound (Unif.) & 0.178 & 0.208 \\
    Lower bound (Human) & 0.031 & 0.033 \\
    \midrule
    Base zero-shot prompt (\texttt{QA}) & 0.173 & 0.206 \\
    Base LLM FT & 0.106 & 0.121  \\
    Chat zero-shot prompt (\texttt{QA}) & 0.308 & 0.383 \\
    Chat LLM FT & 0.109 & 0.148  \\
    \bottomrule
    \end{tabular}
\vspace{5pt}
\end{table}

%% file: tables/unseen_demographics_all.tex
\begin{table*}[ht]
    \centering
    \scriptsize
    \captionsetup{font=small}
    \caption{
    Evaluation performance of our fine-tuned Llama-2-7B model on OpinionQA for subpopulations not included in the fine-tuning dataset \OURDATA-Train.
    For reference, we present a lower bound (human)
    and the zero-shot prompting (\texttt{QA}).
    Absolute difference refers to the WD difference between zero-shot prompting and ours,
    and the relative improvement is calculated in a same way as \Cref{fig:per_group_results}.
    }
    \label{table:unseen_demographic_all}
    \begin{tabular}{cc ccc cc}
    \toprule
      \textbf{Attribute}
    & \textbf{Group}
    & \textbf{Lower Bound (Human)}
    & \textbf{Zero-shot (\texttt{QA})}
    & \textbf{Ours}
    & \textbf{Absolute Diff.}
    & \textbf{Relative Improvement}
    \\
    \midrule
    Age & 18-29 & 0.023 & 0.185 & 0.096 & 0.089 & 0.548 \\
    Age & 30-49 & 0.014 & 0.151 & 0.093 & 0.058 & 0.424 \\
    Age & 50-64 & 0.014 & 0.154 & 0.101 & 0.052 & 0.377 \\
    Age & 65+ & 0.013 & 0.195 & 0.115 & 0.080 & 0.438 \\
    Region & Midwest & 0.016 & 0.153 & 0.095 & 0.058 & 0.425 \\
    Region & West & 0.017 & 0.162 & 0.095 & 0.068 & 0.465 \\
    Education & Associate's Degree & 0.026 & 0.159 & 0.098 & 0.061 & 0.455 \\
    Education & High School Graduate & 0.017 & 0.144 & 0.092 & 0.053 & 0.413 \\
    Education & Postgraduate & 0.015 & 0.174 & 0.106 & 0.068 & 0.426 \\
    Education & Some College, No Degree & 0.018 & 0.144 & 0.093 & 0.051 & 0.405 \\
    Income & \$50,000-\$75,000 & 0.016 & 0.153 & 0.098 & 0.054 & 0.396 \\
    Income & \$30,000-\$50,000 & 0.019 & 0.144 & 0.094 & 0.050 & 0.400 \\
    Political Ideology & Very Conservative & 0.026 & 0.208 & 0.107 & 0.101 & 0.555 \\
    Political Ideology & Very Liberal & 0.025 & 0.202 & 0.111 & 0.091 & 0.514 \\
    Political Party & Independent & 0.016 & 0.155 & 0.093 & 0.062 & 0.445 \\
    Political Party & Something Else & 0.026 & 0.162 & 0.092 & 0.069 & 0.510 \\
    Race & Other & 0.050 & 0.180 & 0.144 & 0.036 & 0.275 \\
    Religion & Agnostic & 0.028 & 0.189 & 0.115 & 0.074 & 0.459 \\
    Religion & Buddhist & 0.063 & 0.207 & 0.149 & 0.059 & 0.405 \\
    Religion & Nothing in Particular & 0.019 & 0.153 & 0.092 & 0.061 & 0.454 \\
    Religion & Orthodox & 0.083 & 0.221 & 0.180 & 0.041 & 0.298 \\
    Religion & Other & 0.051 & 0.184 & 0.123 & 0.061 & 0.457 \\
    Religion & Roman Catholic & 0.018 & 0.145 & 0.098 & 0.047 & 0.371 \\
    \bottomrule    
    \end{tabular}
\end{table*}

%% file: sections/a4_baseline_detail.tex
\section{Baseline Details}
\label{appendix_baseline_detail}

\begin{figure}[!t]
    \captionsetup{font=small}
    \includegraphics[width=\linewidth]{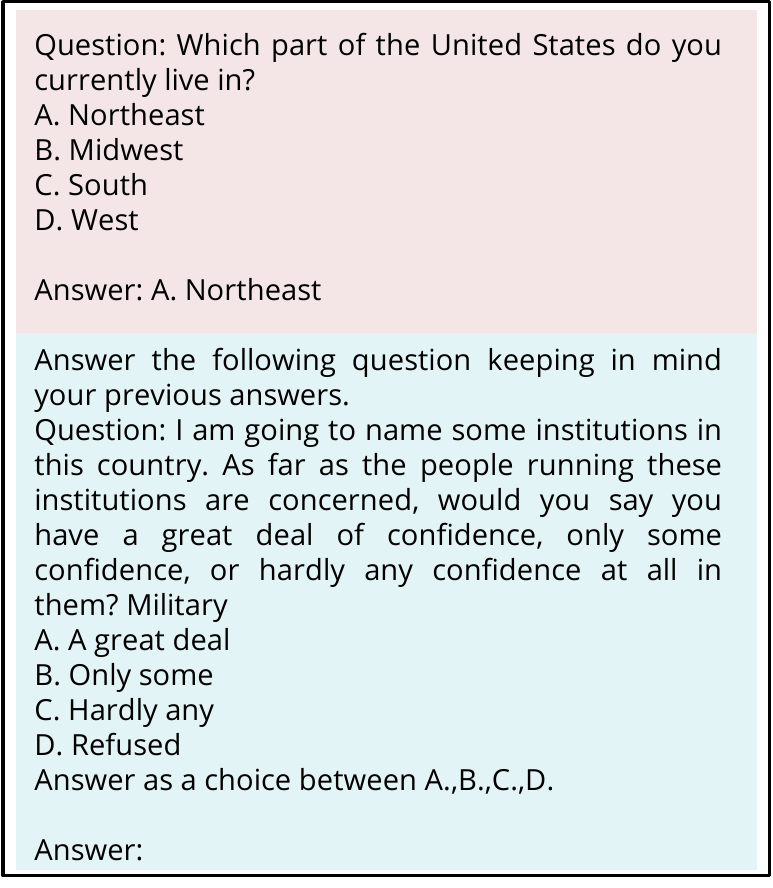}
    \includegraphics[width=\linewidth]{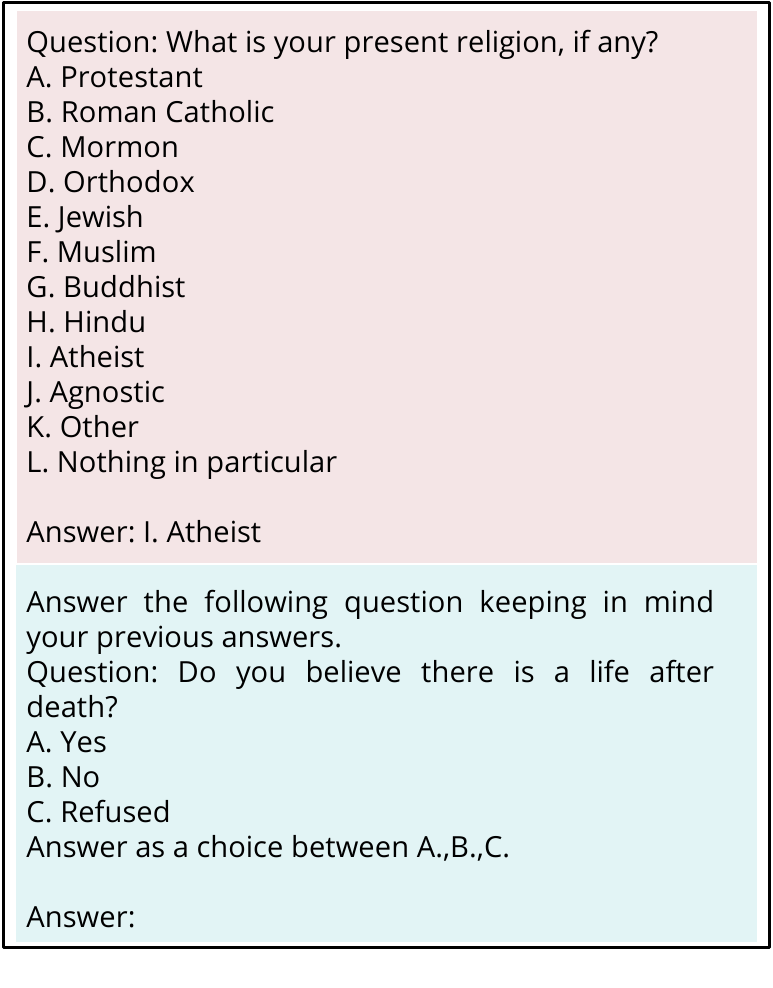}
    \caption{
    Two examples of Zero-shot prompting in the \texttt{QA} format \cite{santurkar2023whose}.
    Subpopulation's information (colored in pink) is concatenated with
    survey question (colored in sky blue).
    The first-token log-probability (probabilities assigned to A, B, C, ...)
    are used to calculate language model's response distribution.
    The same steering prompt format is used in our fine-tuning experiment.
    }
    \label{fig:baseline_qa_example}
\end{figure}

\begin{figure}[!t]
    \captionsetup{font=small}
    \includegraphics[width=\linewidth]{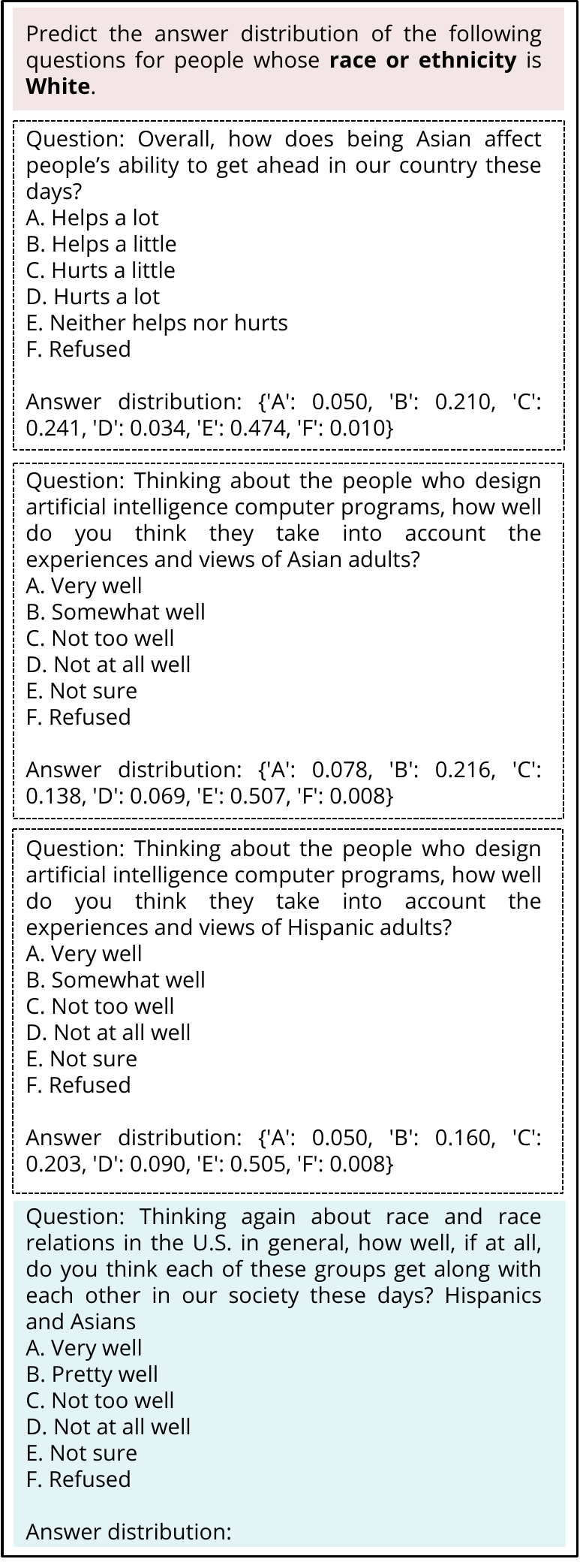}
    \caption{
        Few-shot prompting example for $k=3$.
        Group information is presented in the beginning of the prompt (colored in pink).
        Following group information, $k$ questions whose text embedding are the most similar to the text embedding of the evaluation question (colored in sky blue)
        are presented along with their opinion distribution.
        $k$ questions are presented in the ascending order of cosine similarity.
        The generation of language model (verbalization of opinion distribution)
        is parsed to obtain the response distribution.
    }
    \label{fig:fewshot_example}
\end{figure}

\begin{figure}[!t]
    \captionsetup{font=small}
    \includegraphics[width=\linewidth]{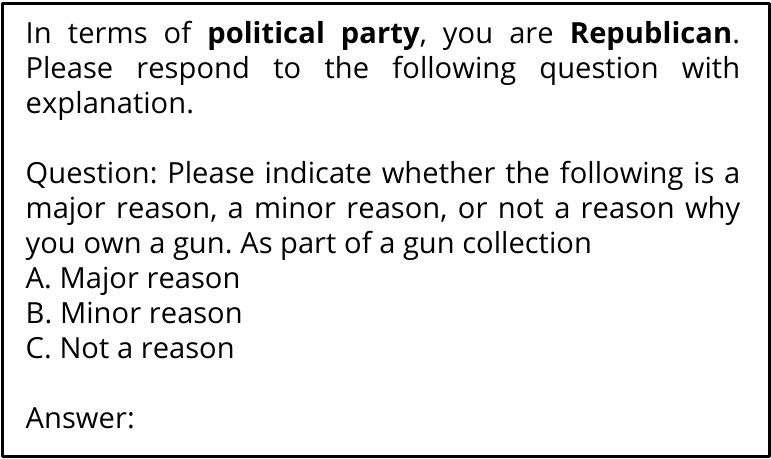}
    \includegraphics[width=\linewidth]{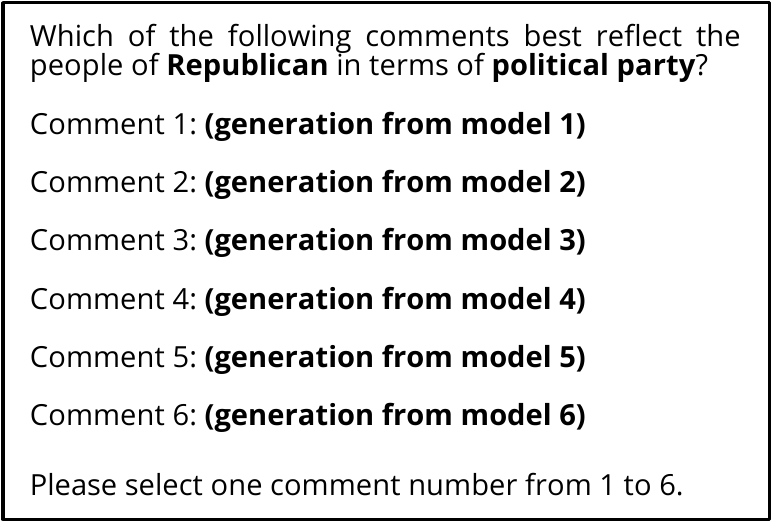}
    \includegraphics[width=\linewidth]{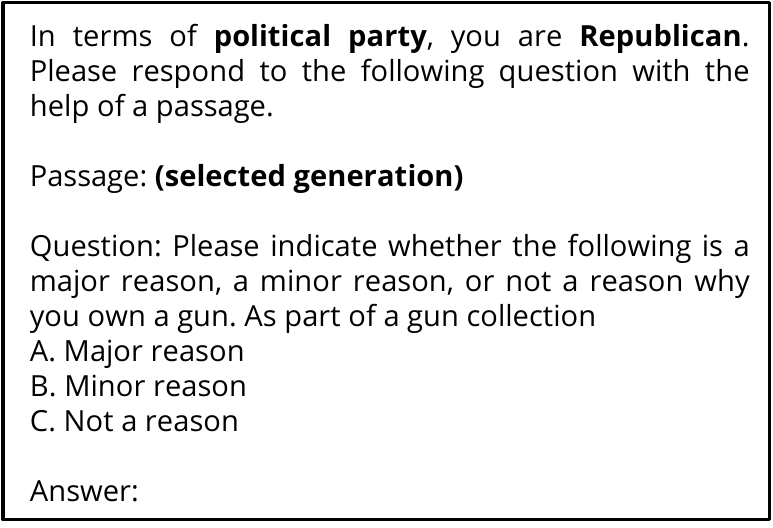}
    \caption{
    Pipeline example of Modular Pluralism.
    Given a demographic group and a survey question, the first prompt is asked to
    multiple (6) language models, Mistral-7B-v0.1-Instruct fine-tuned on the community text corpus.
    The generations are sent to a black-box LLM (gpt-3.5-0613-Instruct) in the format of the second prompt.
    The black-box LLM answers which one of generations best reflects the given demographics.
    Finally, the selected generation serves as a context to answer the given survey question
    and the black-box LLM is prompted (the third prompt) to generate response distribution
    over the answer token A, B, C, etc.
    }
    \label{fig:modular_pluralism_example}
\end{figure}

\begin{itemize}[leftmargin=*]
    \item \textbf{Zero-shot prompting}:
    Three prompt styles—\texttt{QA}, \texttt{BIO}, and \texttt{PORTRAY}—are introduced in~\cite{santurkar2023whose} to integrate group information into prompts.
    These prompts are then combined with survey questions to construct inputs for LLM.
    Then, the first-token log-probability from LLM is measured to calculate the model's response distribution over options.
    In our baseline (and also in fine-tuning experiments) we focus on the \texttt{QA} steering format.
    Examples of this prompting method are shown in \Cref{fig:baseline_qa_example}.
    \item \textbf{Few-shot prompting}: 
    We craft a conditioning prompt that contains not only group information
    but also the group's response distribution to \textit{k} train questions,
    following \cite{hwang2023aligning}.
    For a test question $q_{test} \in Q_{test}$,
    we first sort training questions $Q_{train}$ into $\{q_1, q_2, ...\}$
    such that $\texttt{sim}(\texttt{E}(q_1), \texttt{E}(q_{test}))
    > \texttt{sim}(\texttt{E}(q_2), \texttt{E}(q_{test}))$, and so on.
    $\texttt{E}(q)$ denotes the embedding model (OpenAI-text-embedding-3-large) output
    of the input $q$ and
    \texttt{sim} is a cosine similarity between two embedding vectors. 
    Then, response information of the first $k$ questions $\{q_i, p(\mathcal{A}_{q_i}|q_i, g)\}_{i=1}^{k}$ are used as few shot prompts to have the language model verbalize \cite{meister2024benchmarking} expected response distribution for the given $g$ and $q_{test}$.
    An example of the prompt for $k=3$ case is shown in Figure \ref{fig:fewshot_example},
    while we run the baseline experiment in a $k=5$ setting. 
    
    \item \textbf{Modular Pluralism}:
    The intuition behind Modular Pluralism \cite{feng-etal-2024-modular} is that
    a language model trained on a text corpus of a specific subpopulation will faithfully
    represent public opinion of that population.
    Given a survey question with a \texttt{PORTRAY}-style steering prompt,
    each of language model `modules' (fine-tuned Mistral-7B-Instruct-v0.1)
    generates an option choice with explanation.
    A black-box LLM (GPT-3.5-turbo-Instruct) receives all generations and select a generation that best aligns with the given group. Finally, using the chosen generation as a context, a black-box LLM generates probability distribution over options.
    The example pipeline is shown in Figure \ref{fig:modular_pluralism_example}.
    Instead of the sub-sampled OpinionQA dataset the authors of the method used, we use the exactly same evaluation set across all baseline methods and our approach for a fair comparison.

    \item \textbf{Upper bound}:
    We estimate the distribution between human responses and uniform distribution as an upper bound of WD metrics.
    
    \item \textbf{Lower bound}: We compute a lower bound by randomly sampling a group of respondents and calculating the Wasserstein distance (WD) between the distribution of the sampled group and that of the original respondents for each question. We then bootstrap with $R = 1000$ to construct a 95\% confidence interval (CI) for the WDs. Further details on this estimation process are provided below.

    \paragraph{Computing weighted answer distributions:}
    For each group $g$ and question $q$, we have $n_{gq}$ responses from respondents who belong to group $g$ answering question $q$: $x_1, x_2, \cdots, x_{n_{gq}}$, where $x_i \in \mathcal{A}_q$, i.e., the answer set for question $q$ (e.g., $\{1, 2, 3, 4\}$). 
    Furthermore, each respondent (and thus, their response) is associated with a wave-specific weight $w_1, w_2, \cdots, w_{n_{gq}}$, provided by Pew Research.
    We compute the human answer distribution $\pi_{gq}^{(H)}$ as a weighted sum over responses, where the proportion of respondents providing answer $a \in \mathcal{A}_q$ is estimated as
    \begin{align*}
        \pi_{gq}^{(H)}(a) = \frac{\sum_{i=1}^{n_{gq}} w_i \mathbbm{1}[x_i = a]}{\sum_{i=1}^{n_{gq}} w_i}. \label{eqn:human-weighted-dist}
    \end{align*}

    \paragraph{Bootstrapping at the respondent-level:}
    We draw bootstrap samples per group at the respondent-level including questions from all survey waves.
    This allows us to capture correlations in answer distributions across questions and across waves.
    
    Specifically, let $\mathcal{P}_{g}$ represent the set of respondents in group $g$, where $|\mathcal{P}_{g}| = n_{g}$.
    We produce bootstrapped samples by repeatedly sampling $n_{g}$ respondents from $\mathcal{P}_{g}$ with replacement.
    Let $p_1^{(r)}, p_2^{(r)}, \cdots, p_{n_{g}}^{(r)}$ represent the sampled respondents for the $r$-th bootstrap, and let $w_{1}^{(r)}, w_{2}^{(r)}, \cdots, w_{n_{g}}^{(r)}$ represent their corresponding weights.
    
    For each question $q$, let $\mathcal{P}_{gq} \subseteq \mathcal{P}_{g}$ represent the set of respondents from group $g$ who answered question $q$; as before, $|\mathcal{P}_{gq}| = n_{gq}$.
    Let us define $q(p_i)$ as person $p_i$'s response to question $q$ if $p_i$ answered question $q$, i.e., $p_i \in \mathcal{P}_{gq}$, and 0 otherwise.
    Then, we compute the $r$-th answer distribution to question $q$ as:
    \begin{align*}
        \pi_{gq}^{(r)}(a) = \frac{\sum_{i=1}^{n_{g}} \mathbbm{1}[p_i^{(r)} \in \mathcal{P}_{gq}] w_i^{(r)} \mathbbm{1}[q(p_i^{(r)}) = a]}{\sum_{i=1}^{n_{g}} \mathbbm{1}[p_i^{(r)} \in \mathcal{P}_{gq}] w_i^{(r)}}.
    \end{align*}

    \paragraph{Human lower bound of WD.}
    Our statistic of interest is the mean Wasserstein distance over all questions $Q$ across all waves per group. We approximate this as the WD between the observed human distribution $\pi_{gq}^{(H)}$ and the bootstrap sample $\pi_{gq}^{(r)}$ for question $q$ and group $g$.
    Over all $R=1000$ bootstraps, we have

\begin{align*}
    \mathcal{D}_{g}^{(H)} &= \left\{\frac{1}{|Q|}\sum_{q \in Q} WD(\pi_{gq}^{(H)}, \pi_{gq}^{(r)})\right\}_{r=1}^R.
\end{align*}
To quantify agreement between human samples, we report the mean and 95\% CI (i.e., from $2.5^{th}$ to $97.5^{th}$ percentiles) of $\mathcal{D}_{gq}^{(H)}$.
    
\end{itemize}

%% file: sections/a5_main_result_finegrained.tex
\section{Wave, Group-level Opinion Matching}
\label{appendix_finegrain}
Here we present a group-level and wave-level averaged Wasserstein distance.
Wave-level result is in Table \ref{table:per_wave_opinionqa},
and group-level results for OpinionQA and \OURDATA-Eval are in Table \ref{table:per_group_opinionqa},
\ref{table:per_group_gss}, respectively.
We observe that the improvements in distribution matching between LLM response and human response are consistent across diverse subpopulations and waves.

\input{tables/per_wave_result}
\input{tables/per_group_result}

\input{tables/per_group_result_gss}

%% file: tables/per_wave_result.tex
\begin{table*}[ht]
    \centering
    \scriptsize
    \captionsetup{font=small}
    \caption{
    Per-wave Wasserstein distance on OpinionQA for each base model, comparing baseline zero-shot prompting (\texttt{QA}) with our fine-tuned model Ours(\OURDATA-FT).
    Highlighted rows represent waves whose topics are not covered by the training data (\OURDATA-Train). We observe WD improvement consistently across survey waves and also for waves of topics not covered in the training data.
    }
        \label{table:per_wave_opinionqa}
    \begin{tabular}{c|cc|cc|cc|cc}
    \toprule
    \multirow{2}{*}{\textbf{Wave}}
    & \multicolumn{2}{c|}{Llama-2-7B}
    & \multicolumn{2}{c|}{Llama-2-13B}
    & \multicolumn{2}{c|}{Mistral-7B-v0.1} 
    & \multicolumn{2}{c}{Llama-3-70B} \\
    & Zero-shot & Ours (\OURDATA-FT) & Zero-shot & Ours (\OURDATA-FT) & Zero-shot & Ours (\OURDATA-FT) & Zero-shot & Ours (\OURDATA-FT) \\
    \midrule
    \highlightrowtwo 26 & 0.191 & 0.145 & 0.180 & 0.126 & 0.178 & 0.131 & 0.134 & 0.084 \\
    29 & 0.169 & 0.096 & 0.172 & 0.123 & 0.153 & 0.096 & 0.125 & 0.085 \\
    32 & 0.163 & 0.110 & 0.156 & 0.098 & 0.137 & 0.099 & 0.151 & 0.091 \\
    \highlightrowtwo 34 & 0.155 & 0.105 & 0.171 & 0.089 & 0.134 & 0.095 & 0.138 & 0.083 \\
    36 & 0.175 & 0.120 & 0.184 & 0.126 & 0.175 & 0.107 & 0.130 & 0.087 \\
    \highlightrowtwo 41 & 0.160 & 0.090 & 0.155 & 0.084 & 0.134 & 0.073 & 0.116 & 0.085 \\
    \highlightrowtwo 42 & 0.159 & 0.053 & 0.146 & 0.059 & 0.127 & 0.059 & 0.131 & 0.084 \\
    43 & 0.179 & 0.112 & 0.172 & 0.104 & 0.154 & 0.102 & 0.124 & 0.099 \\
    45 & 0.177 & 0.101 & 0.177 & 0.093 & 0.149 & 0.084 & 0.126 & 0.091 \\
    49 & 0.151 & 0.098 & 0.143 & 0.131 & 0.128 & 0.116 & 0.159 & 0.087 \\
    50 & 0.209 & 0.139 & 0.196 & 0.121 & 0.188 & 0.125 & 0.154 & 0.078 \\
    54 & 0.158 & 0.087 & 0.158 & 0.087 & 0.128 & 0.077 & 0.118 & 0.079 \\
    82 & 0.173 & 0.098 & 0.171 & 0.075 & 0.148 & 0.077 & 0.174 & 0.093 \\
    92 & 0.165 & 0.073 & 0.153 & 0.071 & 0.140 & 0.055 & 0.126 & 0.081 \\
    \bottomrule    
    \end{tabular}
\vspace{1em}
\end{table*}

%% file: tables/per_group_result.tex
\begin{table*}[ht]
    \hspace{-2em}
    \centering
    \scriptsize
    \captionsetup{font=small}
    \caption{
    Per-group Wasserstein distance on OpinionQA for each base models, before and after fine-tuning on \OURDATA-Train. Base refers to zero-shot prompting (\texttt{QA}). *Full group variable name is ``College grad, some Postgrad''.
    }
    \label{table:per_group_opinionqa}
    \begin{tabular}{cc|c|cc|cc|cc|cc}
    \toprule
    \multirow{2}{*}{\textbf{Attribute}}
    & \multirow{2}{*}{\textbf{Group}}
    & \multirow{1}{*}{}
    & \multicolumn{2}{c|}{Llama-2-7B}
    & \multicolumn{2}{c|}{Llama-2-13B}
    & \multicolumn{2}{c|}{Mistral-7B-v0.1} 
    & \multicolumn{2}{c}{Llama-3-70B} \\

    & & Human Baseline & Base & Fine-tuned & Base & Fine-tuned & Base & Fine-tuned & Base & Fine-tuned \\
    \midrule
    \multirow{2}{*}{Region} &
    Northeast & 0.023 & 0.165 & 0.094 & 0.155 & 0.088 & 0.155 & 0.083 & 0.134 & 0.084 \\
    & South & 0.017 & 0.149 & 0.092 & 0.143 & 0.085 & 0.133 & 0.081 & 0.113 & 0.078 \\
    \midrule
    \multirow{2}{*}{Education} &
    College grad* & 0.018 & 0.165 & 0.099 & 0.157 & 0.096 & 0.136 & 0.089 & 0.125 & 0.085 \\
    & Less than high school & 0.043 & 0.161 & 0.101 & 0.150 & 0.096 & 0.134 & 0.094 & 0.151 & 0.091 \\
    \midrule
    \multirow{2}{*}{Gender} &
    Male & 0.015 & 0.182 & 0.093 & 0.152 & 0.089 & 0.131 & 0.083 & 0.138 & 0.083 \\
    & Female & 0.013 & 0.162 & 0.100 & 0.158 & 0.092 & 0.146 & 0.088 & 0.130 & 0.087 \\
    \midrule
    \multirow{4}{*}{Race / ethnicity} &
    Black & 0.031 & 0.151 & 0.102 & 0.144 & 0.095 & 0.132 & 0.091 & 0.116 & 0.085 \\
    & White & 0.012 & 0.176 & 0.097 & 0.178 & 0.093 & 0.145 & 0.085 & 0.131 & 0.084 \\
    & Asian & 0.051 & 0.165 & 0.111 & 0.167 & 0.104 & 0.143 & 0.102 & 0.124 & 0.099 \\
    & Hispanic & 0.044 & 0.162 & 0.102 & 0.163 & 0.098 & 0.134 & 0.092 & 0.126 & 0.091 \\
    \midrule
    \multirow{2}{*}{Income} &
    \$100,000 or more & 0.019 & 0.172 & 0.103 & 0.162 & 0.100 & 0.147 & 0.091 & 0.159 & 0.087 \\
    & Less than \$30,000 & 0.021 & 0.162 & 0.091 & 0.148 & 0.083 & 0.127 & 0.080 & 0.154 & 0.078 \\
    \midrule
    \multirow{2}{*}{Political Party} &
    Democrat & 0.016 & 0.172 & 0.099 & 0.158 & 0.092 & 0.161 & 0.082 & 0.118 & 0.079 \\
    & Republican & 0.019 & 0.196 & 0.105 & 0.235 & 0.101 & 0.181 & 0.095 & 0.174 & 0.093 \\
    \midrule
    \multirow{3}{*}{Political Ideology} &
    Liberal & 0.022 & 0.192 & 0.100 & 0.181 & 0.094 & 0.166 & 0.084 & 0.126 & 0.081 \\
    & Conservative & 0.021 & 0.169 & 0.103 & 0.153 & 0.099 & 0.144 & 0.094 & 0.141 & 0.092 \\
    & Moderate & 0.016 & 0.151 & 0.094 & 0.153 & 0.090 & 0.132 & 0.082 & 0.106 & 0.081 \\
    \midrule
    \multirow{5}{*}{Religion} &
    Protestant & 0.016 & 0.015 & 0.166 & 0.096 & 0.158 & 0.092 & 0.146 & 0.086 & 0.143 \\
    & Jewish & 0.058 & 0.182 & 0.124 & 0.182 & 0.122 & 0.165 & 0.115 & 0.144 & 0.115 \\
    & Hindu & 0.079 & 0.211 & 0.160 & 0.232 & 0.163 & 0.211 & 0.161 & 0.181 & 0.157 \\
    & Atheist & 0.035 & 0.202 & 0.118 & 0.204 & 0.110 & 0.196 & 0.099 & 0.135 & 0.098 \\
    & Muslim & 0.089 & 0.202 & 0.159 & 0.209 & 0.156 & 0.204 & 0.146 & 0.171 & 0.144 \\
    \bottomrule    
    \end{tabular}
\end{table*}

%% file: tables/per_group_result_gss.tex
\begin{table*}[ht]
    \centering
    \scriptsize
    \captionsetup{font=small}
    \caption{
    Per-group Wasserstein distance on \OURDATA-Eval for each base models, before and after fine-tuning on \OURDATA-Train. Base refers to zero-shot prompting (\texttt{QA}). *Full group variable name is ``College grad, some Postgrad''.
    }
    \label{table:per_group_gss}
    \begin{tabular}{cc|c|cc|cc|cc|cc}
    \toprule
    \multirow{2}{*}{\textbf{Attribute}}
    & \multirow{2}{*}{\textbf{Group}}
    & \multirow{1}{*}{}
    & \multicolumn{2}{c|}{Llama-2-7B}
    & \multicolumn{2}{c|}{Llama-2-13B}
    & \multicolumn{2}{c|}{Mistral-7B-v0.1}
    & \multicolumn{2}{c}{Llama-3-70B} \\
    & & Human Baseline & Base & Fine-tuned & Base & Fine-tuned & Base & Fine-tuned & Base & Fine-tuned \\
    \midrule
    \multirow{2}{*}{Region} &
    Northeast & 0.027 & 0.196 & 0.113 & 0.193 & 0.103 & 0.185 & 0.108 & 0.156 & 0.078 \\
    & South & 0.018 & 0.183 & 0.108 & 0.185 & 0.103 & 0.176 & 0.103 & 0.138 & 0.080 \\
    \midrule
    \multirow{2}{*}{Education} &
    College grad* & 0.019 & 0.206 & 0.105 & 0.175 & 0.101 & 0.167 & 0.099 & 0.137 & 0.077 \\
    & Less than high school & 0.036 & 0.191 & 0.129 & 0.182 & 0.117 & 0.172 & 0.121 & 0.180 & 0.108 \\
    \midrule
    \multirow{2}{*}{Gender} &
    Male & 0.017 & 0.186 & 0.102 & 0.176 & 0.101 & 0.170 & 0.099 & 0.150 & 0.079 \\
    & Female & 0.016 & 0.184 & 0.108 & 0.198 & 0.105 & 0.176 & 0.100 & 0.151 & 0.080 \\
    \midrule
    \multirow{4}{*}{Race / ethnicity} &
    Black & 0.029 & 0.200 & 0.114 & 0.179 & 0.102 & 0.170 & 0.107 & 0.139 & 0.094 \\
    & White & 0.014 & 0.190 & 0.105 & 0.187 & 0.103 & 0.181 & 0.102 & 0.153 & 0.083 \\
    & Asian & 0.049 & 0.201 & 0.119 & 0.190 & 0.107 & 0.184 & 0.114 & 0.158 & 0.096 \\
    & Hispanic & 0.050 & 0.204 & 0.133 & 0.199 & 0.122 & 0.182 & 0.134 & 0.172 & 0.115 \\
    \midrule
    \multirow{2}{*}{Income} &
    \$100,000 or more & 0.021 & 0.210 & 0.111 & 0.184 & 0.106 & 0.176 & 0.102 & 0.179 & 0.082 \\
    & Less than \$30,000 & 0.026 & 0.179 & 0.115 & 0.172 & 0.103 & 0.165 & 0.105 & 0.171 & 0.086 \\
    \midrule
    \multirow{2}{*}{Political Party} &
    Democrat & 0.020 & 0.219 & 0.103 & 0.197 & 0.092 & 0.199 & 0.091 & 0.128 & 0.076 \\
    & Republican & 0.023 & 0.205 & 0.123 & 0.234 & 0.117 & 0.206 & 0.115 & 0.187 & 0.093 \\
    \midrule
    \multirow{3}{*}{Political Ideology} &
    Liberal & 0.019 & 0.224 & 0.102 & 0.191 & 0.090 & 0.188 & 0.096 & 0.134 & 0.076 \\
    & Conservative & 0.022 & 0.184 & 0.120 & 0.178 & 0.112 & 0.172 & 0.113 & 0.160 & 0.092 \\
    & Moderate & 0.018 & 0.191 & 0.110 & 0.183 & 0.103 & 0.170 & 0.103 & 0.141 & 0.082 \\
    \midrule
    \multirow{5}{*}{Religion} &
    Protestant & 0.019 & 0.187 & 0.110 & 0.179 & 0.107 & 0.172 & 0.105 & 0.164 & 0.082 \\
    & Jewish & 0.066 & 0.245 & 0.149 & 0.226 & 0.144 & 0.218 & 0.129 & 0.164 & 0.119 \\
    & Hindu & 0.095 & 0.264 & 0.180 & 0.253 & 0.169 & 0.252 & 0.186 & 0.223 & 0.166 \\
    & Atheist & 0.021 & 0.222 & 0.126 & 0.207 & 0.103 & 0.199 & 0.116 & 0.132 & 0.106 \\
    & Muslim & 0.090 & 0.253 & 0.175 & 0.240 & 0.181 & 0.238 & 0.173 & 0.203 & 0.158 \\
    \bottomrule    
    \end{tabular}
\end{table*}